\documentclass{article} 
\usepackage[accepted]{icml2024}

\usepackage{hyperref}
\usepackage{url}
\usepackage[utf8]{inputenc} 
\usepackage[T1]{fontenc}    
\usepackage{hyperref}       
\usepackage{url}            
\usepackage{booktabs}       
\usepackage{amsfonts}       
\usepackage{nicefrac}       
\usepackage{microtype}      
\usepackage{amsmath}
\usepackage{amssymb}
\usepackage{color}
\usepackage{xspace}
\usepackage{dsfont}
\usepackage{caption}
\usepackage{subcaption}

\usepackage{algorithm}
\usepackage{algorithmic}

\usepackage{amsthm}
\usepackage{graphicx}
\usepackage{tabularx}
\usepackage{pifont}
\usepackage{bm}
\usepackage{ulem}
\usepackage{array}
\usepackage{balance}
\usepackage{multirow}
\usepackage{wrapfig}

\usepackage{ulem} 
\usepackage{tikz}
\usepackage{bbding}
\usepackage{cancel}

\newcommand{\ours}{\textit{BayOTIDE }}

\newcommand{\notcheckmark}{{√\surd}\textsuperscript{\textcolor{black}{\kern-0.35em{\bf--}}}}

\newcommand{\icmlFootNote}{\textsuperscript{$\dagger$}:Work done during summer intern at Alibaba Group. \textsuperscript{$\ddagger$}:Work done at Alibaba Group, and now affiliated with Squirrel AI, USA. }

\renewcommand{\d}{{\rm d}}  

\newcommand{\g}{{\bf g}}

\renewcommand{\t}{{\bf t}}
\renewcommand{\u}{{\bf u}}

\newcommand{\x}{{\bf x}}
\newcommand{\y}{{\bf y}}
\newcommand{\z}{{\bf z}}
\newcommand{\A}{{\bf A}}

\newcommand{\F}{{\bf F}}

\renewcommand{\L}{{\bf L}}

\newcommand{\N}{\mathcal{N}}  

\newcommand{\Dcal}{\mathcal{D}}

\renewcommand{\P}{{\bf P}}
\newcommand{\Q}{{\bf Q}}

\newcommand{\bphi}{\boldsymbol{\phi}}

\newcommand{\blambda}{\boldsymbol{\lambda}}

\newcommand{\btheta}{\boldsymbol{\theta}}

\newcommand{\bSigma}{\boldsymbol{\Sigma}}

\newcommand{\bmu}{\boldsymbol{\mu}}

\newcommand{\0}{{\bf 0}}

\newcommand{\ben}{\begin{enumerate}}
\newcommand{\een}{\end{enumerate}}

\newcommand{\EE}{\mathbb{E}}

\newcommand{\cmt}[1]{}

\newcommand{\tp}{{\widetilde{p}}}

\icmltitlerunning{BayOTIDE: Bayesian Online Multivariate Time Series Imputation with Functional Decomposition }

\begin{document}

\twocolumn[
\icmltitle{BayOTIDE: Bayesian Online Multivariate Time Series Imputation \\ with Functional Decomposition }


\icmlsetsymbol{Ali-intern}{$\dagger$}
\icmlsetsymbol{Ali-worked}{$\ddagger$}

\begin{icmlauthorlist}
\icmlauthor{Shikai Fang}{UU,Ali-intern}
\icmlauthor{Qingsong Wen}{Ali,Ali-worked}
\icmlauthor{Yingtao Luo}{CMU,Ali-intern}
\icmlauthor{Shandian Zhe}{UU}
\icmlauthor{Liang Sun}{Ali}

\end{icmlauthorlist}



\icmlaffiliation{CMU}{Carnegie Mellon University, USA}

\icmlaffiliation{UU}{University of Utah, USA}

\icmlaffiliation{Ali}{DAMO Academy, Alibaba Group}


\icmlcorrespondingauthor{Qingsong Wen}{qingsongedu@gmail.com}

\vskip 0.2in
]



\printAffiliationsAndNotice{\icmlFootNote}  


\begin{abstract}
    
    In real-world scenarios such as traffic and energy management, we frequently encounter large volumes of time-series data characterized by missing values, noise, and irregular sampling patterns. While numerous imputation methods have been proposed, the majority tend to operate within a local horizon, which involves dividing long sequences into batches of fixed-length segments for model training. This local horizon often leads to the overlooking of global trends and periodic patterns. More importantly, most methods assume the observations are sampled at regular timestamps, and fail to handle complex irregular sampled time series in various applications. Additionally, most existing methods are learned in an offline manner. Thus, it is not suitable for applications with rapidly arriving streaming data.  To address these challenges, we propose \ours: Bayesian Online Multivariate Time series Imputation with functional decomposition. Our method conceptualizes multivariate time series as the weighted combination of groups of low-rank temporal factors with different patterns.  We employ a suite of Gaussian Processes (GPs),each with a unique kernel, as functional priors to model these factors. For computational efficiency, we further convert the GPs into a state-space prior by constructing an equivalent stochastic differential equation (SDE), and developing a scalable algorithm for online inference. The proposed method can not only handle imputation over arbitrary timestamps, but also offer uncertainty quantification and interpretability for the downstream application. We evaluate our method on both synthetic and real-world datasets. We release the code at \url{https://github.com/xuangu-fang/BayOTIDE}. \end{abstract}
\section{Introduction}

Multivariate time series data are ubiquitous and generated quickly in many real-world applications~\citep{jin2023survey}, such as traffic~\citep{li2015trend} and energy~\citep{zhu2023eforecaster}. However, the collected data are often incomplete and noisy due to sensor failures, communication errors, or other reasons. The missing values in the time series data can lead to inaccurate downstream analysis. Therefore, it is essential to impute the missing values in the time series data in an efficient way. 

Most early methods for time series imputation~\citep{acuna2004treatment,van2011mice,durbin2012time} are based on statistical models. DNN-based imputation methods have gotten boosted attention~\citep{fang2020time} in recent years, for their ability to capture complex non-linear patterns. Another remarkable direction is to apply diffusion models~\citep{song2020score,ho2020denoising} to handle probabilistic imputation, where filling the missing value can be modeled as a denoising process. Most recent work TIDER~\citep{liu2023multivariate} proposed another imputation direction to apply the matrix factorization and decompose the series into disentangled representations.

Despite the success of the proposed methods, they are limited in several aspects. First, most DNN-based and diffusion-based methods are trained by splitting the long sequence into small patches. This local horizon can fail to capture the crucial global patterns~\citep{alcaraz2022diffusion,woo2022cost}, such as trends and periodicity, leading to less interpretability. Second, many methods assume the observations are sampled at regular time stamps, and always under-utilize or ignore the real timestamps. Thus, those models can only impute on fixed-step and discretized time points, instead of the arbitrary time stamps at the whole continuous field. Lastly, in real-world applications such as electricity load monitoring, massive time series data are generated quickly and collected in a streaming manner~\citep{liu2023sadi}. It is extremely costly or even impossible to retrain the model from scratch when new data arrives. Thus, to align with streaming data, the imputation model should work and update in an efficient online manner. However, to the best of our knowledge, all prior imputation methods are designed and optimized in an offline manner, i.e., go through all collected data several epochs for training, which is not suitable for streaming data scenarios.

To handle those limitations, we propose \ours: \underline{Bay}esian \underline{O}nline multivariate \underline{T}ime series \underline{I}mputation with functional \underline{DE}composition. \ours treats the observed values of multivariate time series as the noisy samples from a continuous temporal function. Then, we decompose the function into groups of weighted functional factors, and each factor is aimed to capture a dynamic pattern. We apply Gaussian Processes (GPs) with smooth and periodic kernels as functional priors to fit the factors. Employing the SDE representation of GPs and moment-matching techniques, we develop an online algorithm to infer the running posterior of weights and factors efficiently. As it is a Bayesian model, \ours can offer uncertainty quantification and robustness against noise. The learned functional factors can provide not only interpretability but also imputation over arbitrary timestamps. We list the comparison of \ours and other main-stream imputation methods in Table~\ref{table:compare}. In summary, we highlight our contributions as follows:

\begin{itemize}
    \item We propose \ours, a novel Bayesian method for multivariate time series imputation. \ours can explicitly learn the function factors representing various global patterns, which offer interpretability and uncertainty quantification. As \ours is a continuous model, it can utilize the irregularly sampled timestamps and impute over arbitrary timestamps naturally.

    \item To the best of our knowledge, \ours is the first online probabilistic imputation method of multivariate time series that could fit streaming data well. Furthermore, we develop a scalable inference algorithm with closed-form update and linear cost via moment-matching techniques. 
    
    \item We extensively evaluate our method on synthetic and real-world datasets, and the results show that \ours outperforms the state-of-the-art methods in terms of accuracy and efficiency.
\end{itemize}

\begin{table*}[h]
    \centering
    \begin{small}
        \begin{tabular}{l|ccccc}
            {Properties / Methods } & \textbf{BayOTIDE} & {TIDER} & {Statistic-based} & {DNN-based} & {Diffusion-based} \\
            \midrule
            {Uncertainty-aware} & \ding{51} & \ding{55} & \ding{55} & \bcancel{\Checkmark} & \ding{51} \\
            {Interpretability} & \ding{51} & \ding{51} & \ding{51} & \ding{55} & \ding{55} \\
            {Continuous modeling} & \ding{51} & \ding{55} & \ding{55} &  \ding{55} & \bcancel{\Checkmark} \\
            {Inference manner} & \textbf{online} & \text{offline} & \text{offline} & \text{offline} &\text{offline} \\
        \end{tabular}
    \end{small}
    \caption{\small Comparison of \ours and main-stream multivariate time series imputation methods. \bcancel{\Checkmark} means only partial models in the family have the property, or it's not clear in the original paper. For example, only deep models with probabilistic modules can offer uncertainty quantification, such as GP-VAE~\citep{fortuin2020gp}, but most deep models cannot. The diffusion-based CSDI~\citep{tashiro2021csdi} and CSBI~\citep{chen2023provably} take timestamps as input, but the model is trained with discretized time embedding.}
    \label{table:compare}
\end{table*}

\section{Related Work}

\textbf{Disentangled representations of time series.} The most classical framework of decomposing time series into disentangled representations is the seasonal-trend decomposition(STL)~\citep{cleveland1990stl} along with its following work ~\citep{wen2019robuststl,abdollahi2020novel, bandara2021mstl}, which are non-parametric method to decompose the univariate series into seasonal, trend and residual components. \citep{qiu2018multivariate} proposed the structural design to extend decomposition into multivariate and probabilistic cases. Recently, CoST~\citep{woo2022cost} and TIDER~\citep{liu2023multivariate} show the disentangled representations of multivariate series could get significant performance improvement in forecasting and imputation tasks, respectively, with bonus of interpretability. However, they are not flexible enough to handle the continuous time field and observation noise. \citet{benavoli2021state} propose a similar idea to directly utilize the state-space GPs with mixture kernels to estimate the seasonal-trend factors, but is restricted in univariate series. Besides the imputation and forecasting, learning the disentangled representation of time series is a crucial tool for estimating causal-related latent variables in sequential data \citep{yao2021learning, yao2022temporally}.  A series of recent works on tensor-valued time series analysis \citep{fang2022bayesian,fang2024streaming,wang2024dynamic} also apply the idea of disentangled representations to learn temporal dynamics in latent spaces.   

\textbf{Bayesian imputation modeling.} Bayesian methods are widely used in time series imputation tasks for robust modeling and uncertainty quantification. Early work directly applies a single Bayesian model like Gaussian Process~\citep{roberts2013gaussian} and energy models~\citep{brakel2013training} to model the dynamics. With deep learning boosting in the past ten years, it is popular to utilize the probabilistic modules with various deep networks, such as RNN~\citep{mulyadi2021uncertainty}, VAE~\citep{fortuin2020gp} and GAN~\citep{yoon2018gain}. Adopting score-based generative models (SGMs) is another promising direction for probabilistic imputation, which could be used as autoregressive denoising~\citep{rasul2021autoregressive}, conditional diffusion~\citep{tashiro2021csdi}, Schrödinger Bridge~\citep{chen2023provably} and state-space blocks~\citep{alcaraz2022diffusion}. However, most of the above methods are trained in the offline and patching-sequence manner, which lacks interpretability and may not fit streaming scenarios.

\section{Background}

\subsection{Multivariate Time Series Imputation}
 The classical multivariate time series imputation problem is formulated as follows. A $N$-step multivariate time series $\mathbf{X}=\left\{\mathbf{x}_{1}, \ldots, \mathbf{x}_{N}\right\}$ $ \in \mathbb{R}^{D\times N}$, where $\mathbf{x}_{n} \in \mathbb{R}^{D}$ is the $D$-size value at $n$-th step and $\mathbf{x}_{n}^d$ represents it's values at $d$-th channel. There is a mask matrix $\mathbf{M} \in \{0,1 \}^{ D\times N}$,  indicating whether the series value is observed or missing. The goal is to use the observed values, where $\mathbf{M}_{d,n}=1$, to estimate the missing values $\mathbf{x}_{n}^{d}$, where $\mathbf{M}_{d,n}=0$. In the above setting, the interval between two consecutive timestamps is assumed to be constant by default. If the timestamps are irregularly sampled and continuous, the problem becomes more challenging and the exact timestamps $\{t_{1}, \ldots, t_{N}\}$ should be considered in the imputation model.  In this paper, we aimed to learn a general function $\mathbf{X}(t): t \rightarrow \mathbb{R}^{D}$  to impute the missing values at any time $t \in [t_1, t_N ]$.

\subsection{Gaussian Process (GP) and State-Space Model }\label{ssec:lti-sde}

\textbf{ Gaussian Process (GP)}~\citep{Rasmussen06GP}s is a powerful Bayesian prior for functional approximation, always denoted as $f \sim \mathcal{G} \mathcal{P}\left(0, \kappa\left(\mathbf{x}, \mathbf{x}^{\prime}\right)\right)$.  As a non-parametric model, it's characterized by a mean function, here assumed to be zero, and a covariance function or kernel  $\kappa\left(\mathbf{x}, \mathbf{x}^{\prime}\right)$, which is a positive definite function that measures the similarity of inputs. The choice of the kernel is crucial as it determines the types of functions the GP can model. For instance, the Matérn kernel  
\begin{align}
  & \kappa_{\text{Matérn}}=\sigma^2 \frac{\left(\frac{\sqrt{2 \nu}}{l} \alpha\left(\mathbf{x}, \mathbf{x}^{\prime}\right)\right)^\nu}{\Gamma(\nu) 2^{\nu-1}} K_\nu\left(\frac{\sqrt{2 \nu}}{l} \alpha\left(\mathbf{x}, \mathbf{x}^{\prime}\right)\right), \label{eq:matern-kernel}
\end{align}
and periodic kernel:
\begin{align}
     \kappa_{\text{periodic}}= \sigma^2 exp\left(-2\sin^2(\pi\alpha\left(\mathbf{x}, \mathbf{x}^{\prime}\right)/ p)/l^2\right) \label{eq:periodic-kernel}
 \end{align}

\cmt{
$ \kappa_{\text{Matérn}}=\sigma^2 \frac{\left(\frac{\sqrt{2 \nu}}{l} \alpha\left(\mathbf{x}, \mathbf{x}^{\prime}\right)\right)^\nu}{\Gamma(\nu) 2^{\nu-1}} K_\nu\left(\frac{\sqrt{2 \nu}}{l} \alpha\left(\mathbf{x}, \mathbf{x}^{\prime}\right)\right)$ and periodic kernel: $\kappa_{\text{periodic}}= \sigma^2 exp\left(-2\sin^2(\pi\alpha\left(\mathbf{x}, \mathbf{x}^{\prime}\right)/ p)/l^2\right) $}
are versatile choices to model functions with non-linear and cyclical patterns, respectively.  $\{\sigma^2,l, \nu, p  \}$ are hyperparameters determining the variance, length-scale, smoothness, and periodicity of the function.  $\alpha\left(\cdot, \cdot\right)$ is the Euclidean distance,  and $K_\nu$ is the modified Bessel function,$\Gamma(\cdot)$ is the Gamma function. 

Despite the flexibility and capacity, full GP is a computationally expensive model with $\mathcal{O}(n^3)$ inference cost while handlining $n$  observation data, which is not feasible in practice. To sidestep expensive kernel matrix computation,  \citep{hartikainen2010kalman, sarkka2013bayesian} applied the spectral analysis and worked out a crucial statement: a temporal GP with a stationary kernel is equivalent to a linear time-invariant stochastic differential equation (LTI-SDE). Specifically, given $f(t) \sim \mathcal{G} \mathcal{P}\left(0, \kappa\left(t, t^{\prime}\right)\right)$, we can define a vector-valued companion form: $\mathbf{z}(t) = \left(f(t), \frac{\mathrm{d} f(t)}{\mathrm{d} t}, \ldots, \frac{\mathrm{d} f^m(t)}{\mathrm{d} t}\right)^{\top}: t \rightarrow \mathcal{R}^{m+1}$, where $m$ is the order of the derivative. Then, the GP is equivalent to the solution of the LTI-SDE with canonical form:
\begin{align}
    \frac{\mathrm{d} \mathbf{z}(t)}{\mathrm{d} t}= \mathbf{F} \mathbf{z}(t) + \mathbf{L} {w}(t),   \label{eq:lti-sde}
\end{align}
where $\mathbf{F} $ is a $(m+1) \times (m+1)$ matrix, $\mathbf{L}$ is a $(m+1) \times 1$ vector, and ${w}(t)$ is a white noise process with spectral density $q_\mathbf{s}$.  On arbitrary collection of timestamps $\{ t_1, \ldots, t_N \}$,  the LTI-SDE \eqref{eq:lti-sde} can be further discretized as the Markov model with Gaussian transition, known as the state-space model (SSM):
    \begin{equation}
        \begin{aligned}
          &  p(\mathbf{z}(t_1))= \mathcal{N}(\mathbf{z}(t_1) | \mathbf{0}, \mathbf{P_{\infty}}),   \\ 
           & p(\mathbf{z}(t_{n+1}) | \mathbf{z}(t_{n})) = \mathcal{N}(\mathbf{z}(t_{n+1}) | \mathbf{A}_n \mathbf{z}(t_{n}),\mathbf{Q}_n ), \label{eq:ssm}
        \end{aligned}
    \end{equation}
where $\mathbf{A}_n =  \operatorname{exp}(\mathbf{F}\Delta_n)$, $\mathbf{Q}_n = \int_{t_n}^{t_{n+1}} \mathbf{A}_n \mathbf{L} \mathbf{L}^{\top} \mathbf{A}_n^{\top} q_\mathbf{s} \mathrm{d} t$, $\Delta_n = t_{n+1} - t_n$, and $\mathbf{P_{\infty}}$ is the steady-state covariance matrix, which can be obtained by solving the Lyapunov equation $\mathbf{F} \mathbf{P_{\infty}} + \mathbf{P_{\infty}} \mathbf{F}^{\top} + \mathbf{L} \mathbf{L}^{\top} q_\mathbf{s}= 0$~\citep{lancaster1995algebraic}. SSM \eqref{eq:ssm} can be efficiently solved with linear cost by classical methods like the Kalman filter~\citep{kalman1960new}. After inference over $\mathbf{z}(t)$, we can easily obtain $f(t)$ by a simple projection: $f(t) = [1, 0, \ldots 0]  \mathbf{z}(t)$.

We highlight that all the parameters of the LTI-SDE  \eqref{eq:lti-sde} and its SSM \eqref{eq:ssm}: $\{m, \mathbf{F}, \mathbf{L}, q_\mathbf{s},\mathbf{P_{\infty}}\}$  are time-invariant constant and can be derived from the given stationary kernel function. In practice, stationary kernels are a common choice for GP, which requires the kernel to be a function of the distance between two inputs. For example, the Matérn and periodic kernels are stationary kernels, and we can work out their closed-form formulas of LTI-SDE and SSM. We omit the specific formulas here and refer the readers to the appendix.

\section{Method}
\subsection{Functional Decomposition of Time Series}
The motivation of \ours is based on the fact that the different channels of real-world multivariate time series are always correlated, and there may exist shared patterns across channels. Thus, we propose to decompose the series into a set of functional basis (factors) and channel-wise weights. Inspired by the classic seasonal-trend decomposition~\citep{cleveland1990stl} and TIDER~\citep{liu2023multivariate}, we assume there are two groups of factors representing different temporal patterns. The first group of factors is supposed to capture the nonlinear and long-term patterns,  and the second represents the periodic parts, namely, trends and seasonalities. Thus, we decompose the function $\mathbf{X}(t): t \rightarrow \mathbb{R}^{D}$ as the weighted combination of two groups of functional factors. Specifically,  assume there are $D_{r}$ trend factors and $D_{s}$ seasonality factors, then we have the following decomposition:
\begin{align}
    \mathbf{X}(t) = \mathbf{U}\mathbf{V}(t) =[\mathbf{U}_{\text{trend}}, \mathbf{U}_{\text{season}}]  \left[\begin{array}{c}
        \mathbf{v}_{\text{trend}}(t), \\
        \mathbf{v}_{\text{season}}(t)
        \end{array}\right], \label{eq:decomposition}
\end{align}
where $\mathbf{U} = [\mathbf{U}_{\text{trend}} \in \mathbb{R}^{D \times D_{r}},\mathbf{U}_{\text{season}} \in \mathbb{R}^{D \times D_{s}} ]$  are the weights of the combination. The trends factor group $\mathbf{v}_{\text{season}}(t): t \rightarrow \mathbb{R}^{D_{s}}, $ and seasonality factor groups $\mathbf{v}_{\text{trend}}(t): t \rightarrow \mathbb{R}^{D_{r}} $  are the concatenation of  independent temporal factors over each dimension:
\begin{equation}
    \begin{aligned}
        &\mathbf{v}_{\text{trend}}(t) = \operatorname{concat}[{v}_{\text{trend}}^{i}(t)]_{i=1 \ldots D_r}, \\
        & \mathbf{v}_{\text{season}}(t) = \operatorname{concat}[{v}_{\text{season}}^{j}(t)]_{j=1 \ldots D_s},   
    \end{aligned}
\end{equation}
where ${v}_{\text{trend}}^{i}(t), {v}_{\text{season}}^{j}(t)$ are the 1-d temporal function factor, aimed to model the $i$-th trend and $j$-th seasonality latent pattern respectively.

For the imputation task, if we can estimate the $\mathbf{U}$ and  $\mathbf{V}(t)$ well, we can impute the missing values of $\mathbf{X}(t)$ by $\mathbf{U}\mathbf{V}(t)$ for any $t$. As TIDER~\citep{liu2023multivariate} proposed a low-rank decomposition similar to \eqref{eq:decomposition}, our model can be seen as a generalization of TIDER to the continuous-time and functional field with the Bayesian framework.

\subsection{ GP Prior and Joint Probability of Our Model}

We assume $\mathbf{X}(t)$ is partially observed with missing values and noise on timestamps $ \{t_1, \ldots t_N\}$. We denote the observed values as $\mathbf{Y}= \{\mathbf{y}_n  \}_{n=1}^{N} $, where $\mathbf{y}_n \in \mathbb{R}^D $, and its value at $d$-th channel is denoted as ${y}_n^d$. $\mathbf{M} \in \{0,1\}^{D \times N}$ is the mask matrix, where $1$ for observed values and $0$ for missing values. The noise level is assumed to be the same for all the channels. Thus, we set the Gaussian likelihood for the observed values as:
\begin{align}
    p(\mathbf{Y}|\mathbf{U},\mathbf{V}(t), \tau) = \prod_{(d,n) \in \Omega } \mathcal{N}({y}_{n}^{d} \mid \mathbf{u}^{d} \mathbf{v}(t_n)  , \tau^{-1}),  \label{eq:likelihood}
\end{align}
where $\tau$ is the inverse of the noise level. $\Omega$ is the collection of observed values' location, namely $\Omega= \{(d,n) \mid \mathbf{M}_{d,n}=1\}$. $\mathbf{u}^{d} \in \mathbb{R}^{1\times (D_r + D_s )}$ is the $d$-th row of $\mathbf{U}$, and $\mathbf{v}(t_n) \in \mathbb{R}^{ (D_r + D_s ) \times 1}$ is the concatenation of $\mathbf{v}_{\text{trend}}(t_n)$ and $\mathbf{v}_{\text{season}}(t_n)$.

As $\mathbf{v}_{\text{season}}(t)$ and $\mathbf{v}_{\text{trend}}(t)$ are supposed to capture different temporal patterns, we adopt Gaussian Processes (GPs) with different kernels to model them. Specifically, we use the Matérn kernel to model the trend factors, and the periodic kernel to model the seasonality factors: 
\begin{equation}
    \begin{aligned}
        {v}_{\text{trend}}^{i}(t) \sim \mathcal{GP} (0, \kappa_{\text{Matérn}}),  
       {v}_{\text{season}}^{j}(t) \sim \mathcal{GP} (0, \kappa_{\text{periodic}}). \label{eq:gp-prior}
    \end{aligned}
\end{equation}

We further assume that Gaussian prior for $\mathbf{u}^{d}$ and Gamma prior for $\tau$. Then, the joint probability model is:
\begin{align}
    &p(\mathbf{Y},\mathbf{V}(t),\mathbf{U},\tau ) = \operatorname{Gam}(\tau\mid a_0, b_0) \prod_{d=1}^{D}\mathcal{N}(\mathbf{u}^{d} \mid \mathbf{0}, \mathbf{I}) \nonumber \\
    & \cdot \prod_{j=1}^{D_s}  \mathcal{GP} (0, \kappa_{\text{periodic}}) \prod_{i=1}^{D_r}  \mathcal{GP} (0, \kappa_{\text{Matérn}})\cdot p(\mathbf{Y}| \Theta ),  \label{eq:joint-prob}   
\end{align}
where $\Theta = \{ \mathbf{U}, \mathbf{V}(t), \tau\}$ denotes all model random variables for compactness. As each GP prior term corresponds to a companion form $\mathbf{z}(t)$ (see section\eqref{ssec:lti-sde}), we define the concatenation of all factors' companion forms as $\mathbf{Z}(t)$ and we have $p(\mathbf{V}(t)) = p(\mathbf{Z}(t)) = P(\mathbf{Z}(t_1)) \prod_{i=1}^{N-1}P(\mathbf{Z}(t_{n+1})|\mathbf{Z}(t_{n}))$. 

\subsection{Online Inference}
With the joint probability \eqref{eq:joint-prob}, we further propose an online inference algorithm to estimate the running posterior of $ \Theta$. We denote all observations up to time $t_n$ as $\mathcal{D}_{t_n}$, i.e. $\mathcal{D}_{t_n} = \{ \mathbf{y}_{1}, \ldots, \mathbf{y}_{n}\}$. When a new observation $\mathbf{y}_{n+1}$ arrives at $t_{n+1}$, we aimed to update the posterior distribution $p( \Theta \mid \mathcal{D}_{t_n} \cup \mathbf{y}_{n+1})$ without reusing the previous observations $\mathcal{D}_{t_n}$.  The general principle for online inference is the incremental version of Bayes'rule, which is:
\begin{align}
    p(\Theta \mid \mathcal{D}_{t_n} \cup \mathbf{y}_{n+1}) \propto p(\mathbf{y}_{n+1} \mid \Theta, \mathcal{D}_{t_n}) p(\Theta \mid \mathcal{D}_{t_n}). \label{eq:online-inference}
\end{align}
However, the exact posterior is not tractable. Thus, we first apply the mean-field factorization to approximate the posterior. Specifically, we approximate the exact posterior as: \cmt{ $ p(\Theta \mid \mathcal{D}_{t_n}) \approx q(\Theta \mid \mathcal{D}_{t_n}) = q(\tau \mid \mathcal{D}_{t_n}) \prod_{d=1}^{D}q(\mathbf{u}^{d} \mid \mathcal{D}_{t_n}) q(\mathbf{Z}(t) \mid \mathcal{D}_{t_n})$}
\begin{align}
    p(\Theta \mid \mathcal{D}_{t_n}) \approx  q(\tau \mid \mathcal{D}_{t_n}) \prod_{d=1}^{D}q(\mathbf{u}^{d} \mid \mathcal{D}_{t_n}) q(\mathbf{Z}(t) \mid \mathcal{D}_{t_n}), \label{eq:mean-field}
\end{align}
where $q(\mathbf{u}^d \mid \mathcal{D}_{t_n}) = \N(\mathbf{m}_n^d,\mathbf{V}_n^d)$  and $q(\tau \mid \mathcal{D}_{t_n}) =  \operatorname{Gamma}(\tau\mid a_n, b_n)$ are the approx. distributions for $\mathbf{U}$ and $\tau$ at time $t_n$ respectively. For $\mathbf{Z}(t)$, we design the approximated posterior as $q(\mathbf{Z}(t)\mid \mathcal{D}_{t_n}) = \prod_{i=1}^{n} q(\mathbf{Z}(t_i))$, where $q(\mathbf{Z}(t_i))$ are the concatenation of $q(\mathbf{z}(t_i))= \N(\mathbf{\mu}_{i},\mathbf{S}_{i})$ across all factors. $\{\{\mathbf{m}_n^d,\mathbf{V}_n^d\},  \{\mathbf{\mu}_{i},\mathbf{S}_{i} \}, a_n, b_n, \}$ are the variational parameters of the approximated posterior to be estimated.

With mean-field approximation, \eqref{eq:online-inference} is still not feasible. It's because the multiple factors and weights interweave in the likelihood, and R.H.S of  \eqref{eq:online-inference} is unnormalized. To solve it, we propose a novel online approach to update $q(\Theta \mid \mathcal{D}_{t_n})$  with a closed form by adopting conditional Expectation Propagation(CEP)~\citep{wang2019conditional} and chain structure of $\mathbf{Z}(t)$. Specifically, with current $q(\Theta \mid \mathcal{D}_{t_n})$ and prior $p(\Theta)$, we can approximate each likelihood term of the new-arriving observation ${y}_{n+1}$ as several messages factors:
\begin{align}
    p({y}_{n+1}^{d} \mid \Theta) \approx \mathcal{Z}  f_{n+1}^{d}(\mathbf{Z}(t_{n+1}))f_{n+1}^{d}(\u_d)f_{n+1}^{d}(\tau), \label{eq:approx-likelihood}
\end{align}
where $ \mathcal{Z}$ is the normalization constant, $f_{n+1}^{d}(\u_d) = \mathcal{N}(\u_d \mid \mathbf{\hat{m}}_{n+1}^{d}, \mathbf{\hat{V}}_{n+1}^{d})$ and $f_{n+1}^{d}(\tau) = \operatorname{Gamma}(\tau \mid \hat{a}_{n+1},\hat{ b}_{n+1})$,  $f_{n+1}^{d}(\mathbf{Z}(t_{n+1})) = \operatorname{concat}[\mathcal{N}(\hat{\mu}_i, \hat{{S}_i})] $ are the approximated message factors of $\u_d$, $\tau$, and $\mathbf{Z}(t_{n+1})$ respectively. 

Then, we merge all the message factors of $\mathbf{U}$ and $\tau$ and follow the variational form of \eqref{eq:online-inference}, and will obtain the update equations:
\begin{align}
    &q(\tau | \mathcal{D}_{t_{n+1}})=q(\tau | \mathcal{D}_{t_{n}})\prod_{d=1}^{D}f_{n+1}^{d}(\tau), \label{eq:update-tau} \\
    &q(\mathbf{{u}}^d | \mathcal{D}_{t_{n+1}}) = q(\mathbf{{u}}^d | \mathcal{D}_{t_{n}}) f_{n+1}^{d}(\mathbf{u}^d). \label{eq:update-U}
\end{align}
\cmt{As the R.H.S of \eqref{eq:update-tau} and \eqref{eq:update-U} are all in the same distribution family in the exponential family, we can obtain the closed-form update for $q(\tau | \mathcal{D}_{t_{n+1}})$ and $q(\mathbf{{u}}^d | \mathcal{D}_{t_{n+1}})$.} 

The message approximation \eqref{eq:approx-likelihood} and the message merging procedure \eqref{eq:update-tau}\eqref{eq:update-U}  are based on the conditional moment-matching technique, which can be done in parallel for all channels with closed-form update. We omit the derivation and exact formulas and refer the readers to the appendix. 

Then, we present the online update of $\mathbf{Z}(t)$. With the chain structure of $q(\mathbf{Z}(t)\mid \mathcal{D}_{t_n})$ and $p(\mathbf{Z}(t))$, we found the update can be conducted sequentially. Specifically:
\begin{align}
    &q(\mathbf{Z}(t_{n+1})) = \nonumber \\
    &q(\mathbf{Z}(t_{n})) p(\mathbf{Z}(t_{n+1})|\mathbf{Z}(t_{n})) \prod_{d=1}f_{n+1}^{d}(\mathbf{Z}(t_{n+1})),\label{eq:update-Z}
\end{align}
where $p(\mathbf{Z}(t_{n+1})|\mathbf{Z}(t_{n}))$ is the concatenation of all factors' transition given in \eqref{eq:ssm}. If we treat $ \prod_{d}f_{n+1}^{d}(\mathbf{Z}(t_{n+1}))$ as the observation of the state space, \eqref{eq:update-Z} is the Kalman filter model~\citep{kalman1960new}. Thus, we can obtain the closed-form update of $q(\mathbf{Z}(t)|\mid \mathcal{D}_{t_n} )$, which is the running posterior $q(\mathbf{Z}_{n}|\mathbf{y}_{1:n})$. After going through all observations, we run Rauch-Tung-Striebel (RTS) smoother~\citep{rauch1965maximum} to efficiently compute the full posterior of each state $q(\mathbf{Z}_{n}|\mathbf{y}_{1:N})$ from backward.

The online algorithm is summarized in Algorithm table \ref{alg:alg}: we go through all the timestamps, approximate the message factors with moment-matching, and run Kalman filter and message merging and update sequentially. For each timestamp, we can run moment-matching and posterior update steps iteratively several times with damping trick~\citep{minka2001expectation} for better approximation. The algorithm is very efficient as the message approximation \eqref{eq:approx-likelihood} can be parallel for different channels, and all the updates are closed-form. The algorithm is with time cost $\mathcal{O}(N(D_s + D_r))$ and space cost $\mathcal{O}( \sum_{k=1}^{D_r+D_s} N (m_k+ m_{k}^2) + D (D_s + D_r) )$ where $N$ is the number of timestamps, $D$ is the number of channel of original time series, $D_r$, $D_s$ are number of trends and seasonality factors respectively, $m_k$ is the order of the companion form of $k$-th factor's GP prior determined by the kernel types.

\subsection{
    Probabilistic Imputation at Arbitrary Timestamp}
 With the current posterior $\{q(\mathbf{z}(t_1))\ldots q(\mathbf{z}(t_N))\}$ over the observed timestamps, the functional and chain property of GP priors allow us to infer the prediction distribution, namely do probabilistic interpolation at arbitrary time stamps. Specifically, for a never-seen timestamp $t^{\star}\in (t_1,\t_N )$, we can identify the closest neighbor of $t^{\star}$ observed in training, i.e, $t_k < t^{\star} < t_{k+1}$, where $t_k, t_{k+1} \in \{t_1 \ldots t_N\}$. Then the predictive distribution at $t^{\star}$is given by $q(\mathbf{z}(t^{\star})) = \mathcal{N}(\mathbf{z}^{\star} \mid \mathbf{m}^{\star},\mathbf{V}^{\star})$. $\{\mathbf{m}^{\star},\mathbf{V}^{\star}\}$ are defined as:
 \begin{equation}
    \begin{aligned}
        &\mathbf{V}^{\star} = ( \mathcal{Q}_1^{-1} + \mathcal{A}_2^{\top} \mathcal{Q}_1^{-1} \mathcal{A}_2 )^{-1}, \\
         &\mathbf{m}^{\star} = \mathbf{V}^{\star}(\mathcal{Q}_1^{-1} \mathcal{A}_1 \mathbf{m}_k +\mathcal{A}_2^{\top} \mathcal{Q}_2^{-1}\mathbf{m}_{k+1} ),  \label{eq:predictive}
       \end{aligned}
 \end{equation}
where $\mathbf{m}_k, \mathbf{m}_{k+1}$ are the predictive mean of $q(\mathbf{z}(t_k)),q(\mathbf{z}(t_{k+1}))$, $\{\mathcal{A}_1, \mathcal{A}_2, \mathcal{Q}_1, \mathcal{Q}_2\}$ are transition matrices and covariance matrices based on the forward-backward transitions $p(\mathbf{z}(t^{\star})|\mathbf{z}(t_k)),p(\mathbf{z}(t_{k+1})|\mathbf{z}(t^{\star}))$ respectively. Eq.\eqref{eq:predictive} offers continuous modeling of the time series. We give a detailed derivation in the appendix.

\begin{algorithm}[tb]
	\caption{\ours}
	\label{alg:alg}
	\begin{algorithmic}
		\STATE {\bfseries Input:} observation $\mathbf{Y}= \{\mathbf{y}_n  \}_{n=1}^{N} $over $\{t_n  \}_{n=1}^{N}$, $D_s, D_r$, the kernel hyperparameters.
        \STATE Initialize $q(\tau), q(\mathcal{W}), \{q(\mathbf{Z}(t_n))\}_{n=1}^{N}$.
		\FOR{$t=1$ {\bfseries to} $N$}	
		\STATE Approximate messages by \eqref{eq:approx-likelihood} for all observed channels in parallel.
		\STATE  Update posterior of $\tau$ and $\mathbf{U}$ by \eqref{eq:update-tau} and \eqref{eq:update-U} for all observed channels in parallel.
        \STATE  Update posteriors of $\mathbf{Z}(t)$ using Kalman filter by \eqref{eq:update-Z}.
		\ENDFOR
		\STATE Run RTS smoother to obtain the full posterior of $\mathbf{Z}(t)$.
		\STATE {\bfseries Return:}  $q(\tau), q(\mathcal{W}), \{q(\mathbf{Z}(t_n))\}_{n=1}^{N}$
	\end{algorithmic}
\end{algorithm}

\section{Experiments}

\subsection{Synthetic data}
We first evaluate \ours on a synthetic task. We first set four temporal functions and a weight matrix, defined as follows:
\begin{align}
    &\mathbf{U}=\left(\begin{array}{cccc}
        1 & 1 & -2 & -2 \\
         0.4& 1 & 2 & -1 \\
         -0.3 & 2 & 1 & -1 \\ 
         -1 & 1 &1 & 0.5
        \end{array}\right), \nonumber \\
    & \mathbf{V}(t) = \left(\begin{array}{c}
        10t, \\ 
        \sin(20\pi t), \\
        \cos(40\pi t), \\ 
        \sin(60\pi t)
        \end{array}\right). \nonumber
\end{align}
Then, the four-channel time series is defined as $\mathbf{X}(t) = \mathbf{U}  \mathbf{V}(t)$, and each channel is a mixture of multiscale trend and seasonality factors. We collected 2000 data points over the 500 irregularly sampled timestamps from $[0,1]$. We randomly set only $20 \%$ of the data as observed values, and the rest as missing for evaluation. We further add Gaussian noise with a standard deviation $0.1$ to the observed data. We use the Matérn kernel with $\nu=3 / 2$ as the trend kernel and the periodic kernel with period $20\pi$ as the seasonality kernel. We set $D_r = 1, D_s = 3$.  We highlight that evaluation could be taken on the timestamps that 
never appear in the training set, known as the all-channel-missing imputation, but we can apply \eqref{eq:predictive} to handle such hard cases easily. 

\begin{figure*}
		\centering
        \begin{subfigure}[b]{0.90\textwidth}
            \centering
            \includegraphics[width=\linewidth]{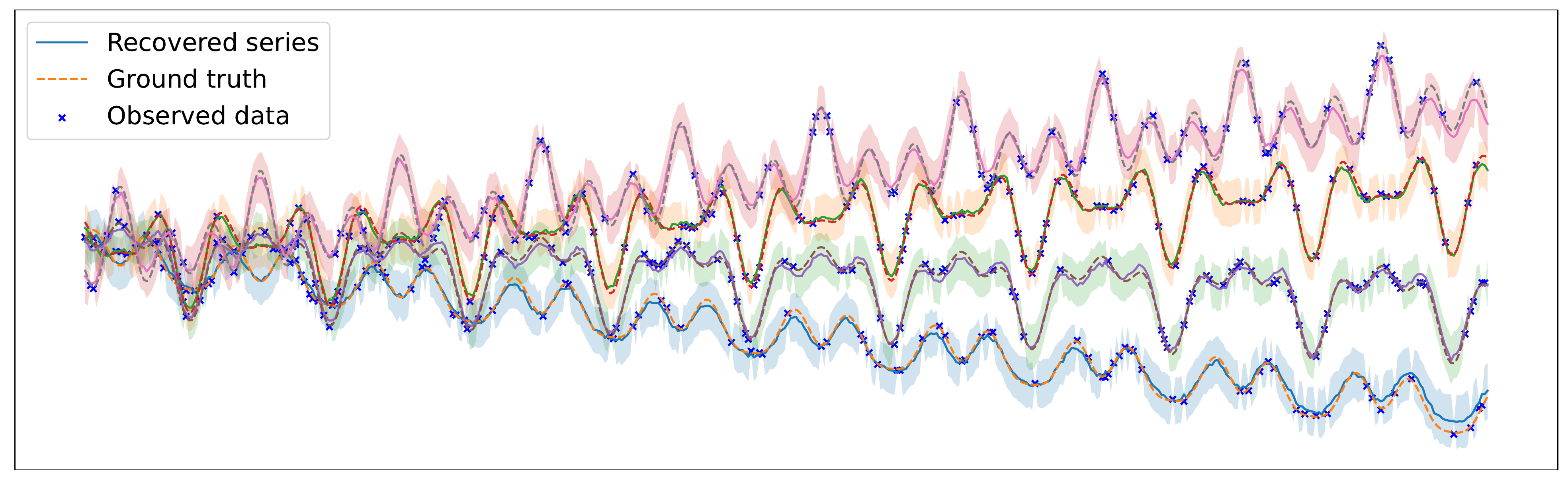}
            \caption{\small Imputation results of the four-channel synthetic time series.}
            \label{fig:simu-impute}
        \end{subfigure}
        \vspace{-0.2in}    
		\setlength{\tabcolsep}{0pt}
		\begin{tabular}[c]{cccc}
			\begin{subfigure}[b]{0.22\textwidth}
				\centering
				\includegraphics[width=\linewidth]{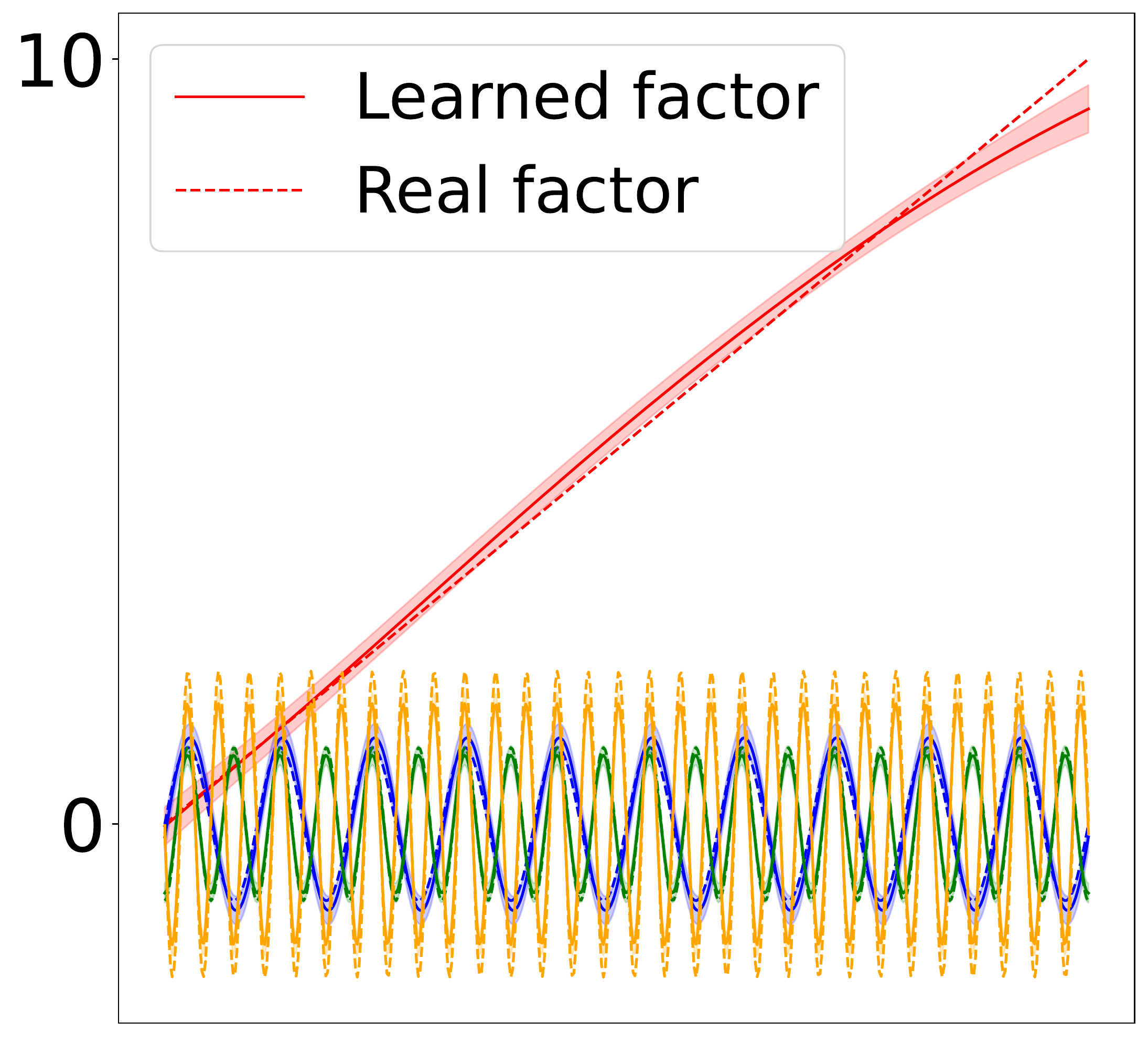}
                \caption{\small Channel\#1's factors}
				\label{fig:simu-channel-1}
			\end{subfigure} &
			\begin{subfigure}[b]{0.215\textwidth}
				\centering
				\includegraphics[width=\linewidth]{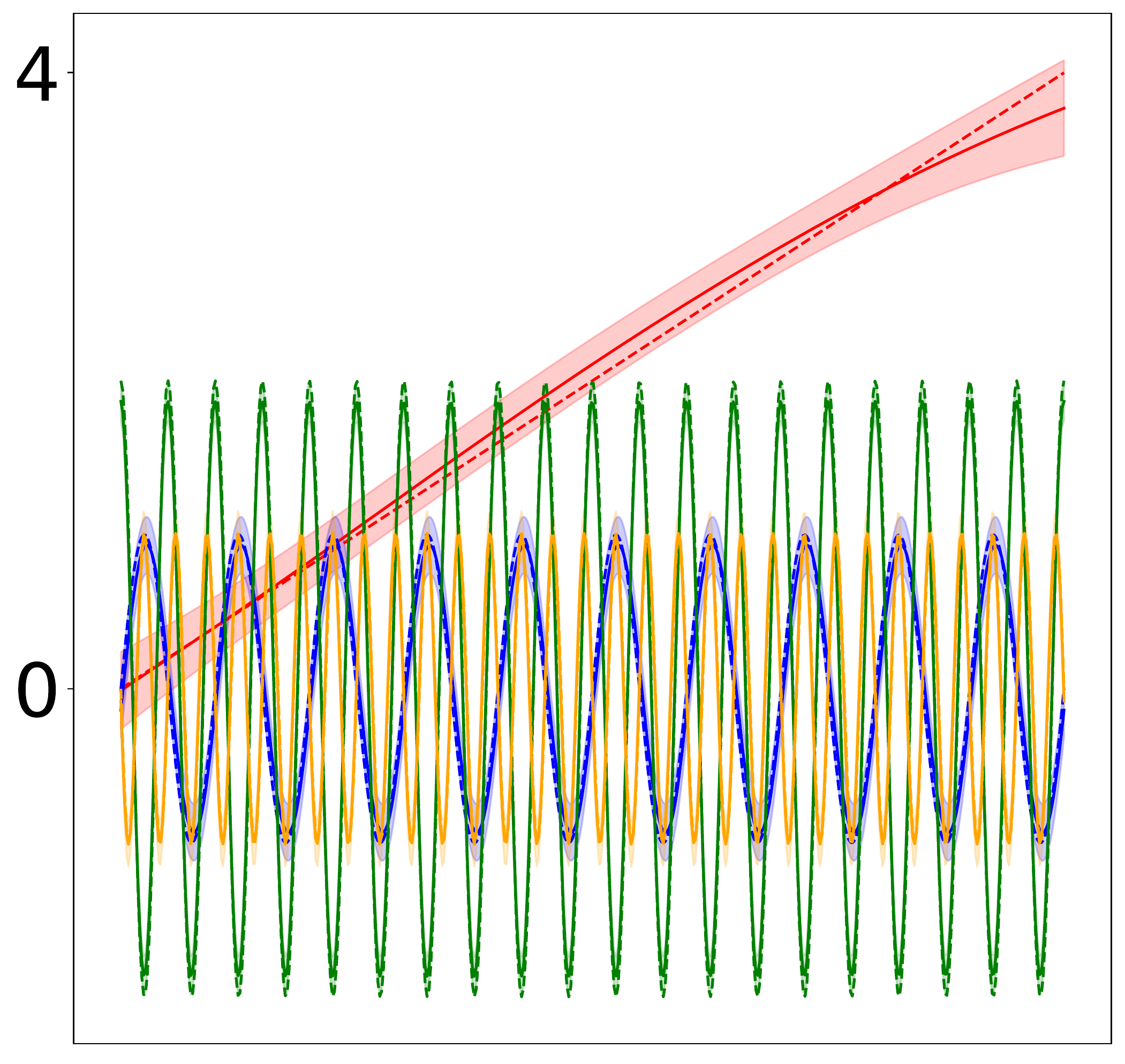}
                \caption{\small Channel\#2's factors}
                \label{fig:simu-channel-2}
			\end{subfigure} &
		\begin{subfigure}[b]{0.222\textwidth}
			\centering
			\includegraphics[width=\linewidth]{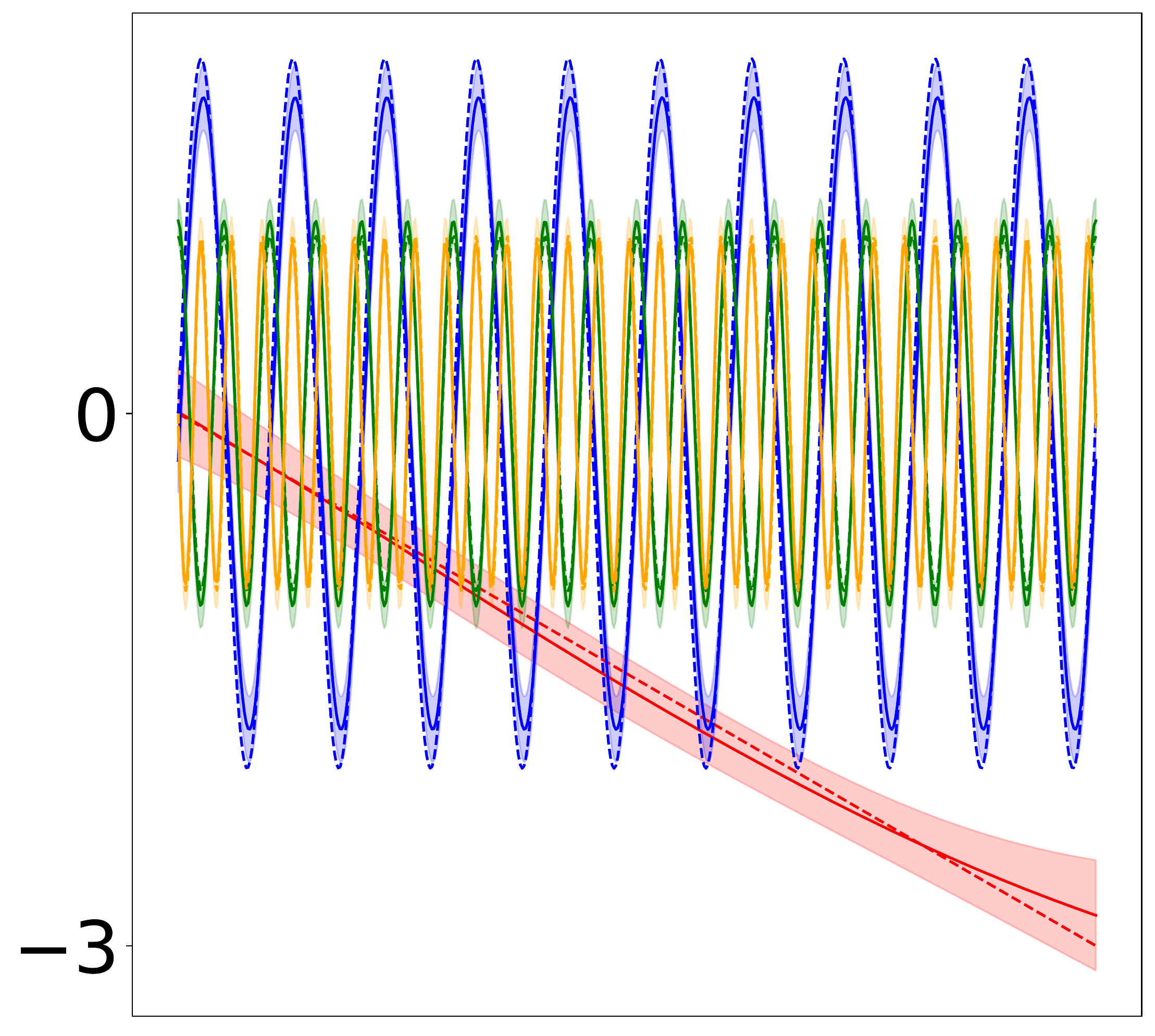}
            \caption{\small Channel\#3's factors}
            \label{fig:simu-channel-3}
		\end{subfigure} &
		\begin{subfigure}[b]{0.222\textwidth}
			\centering
			\includegraphics[width=\linewidth]{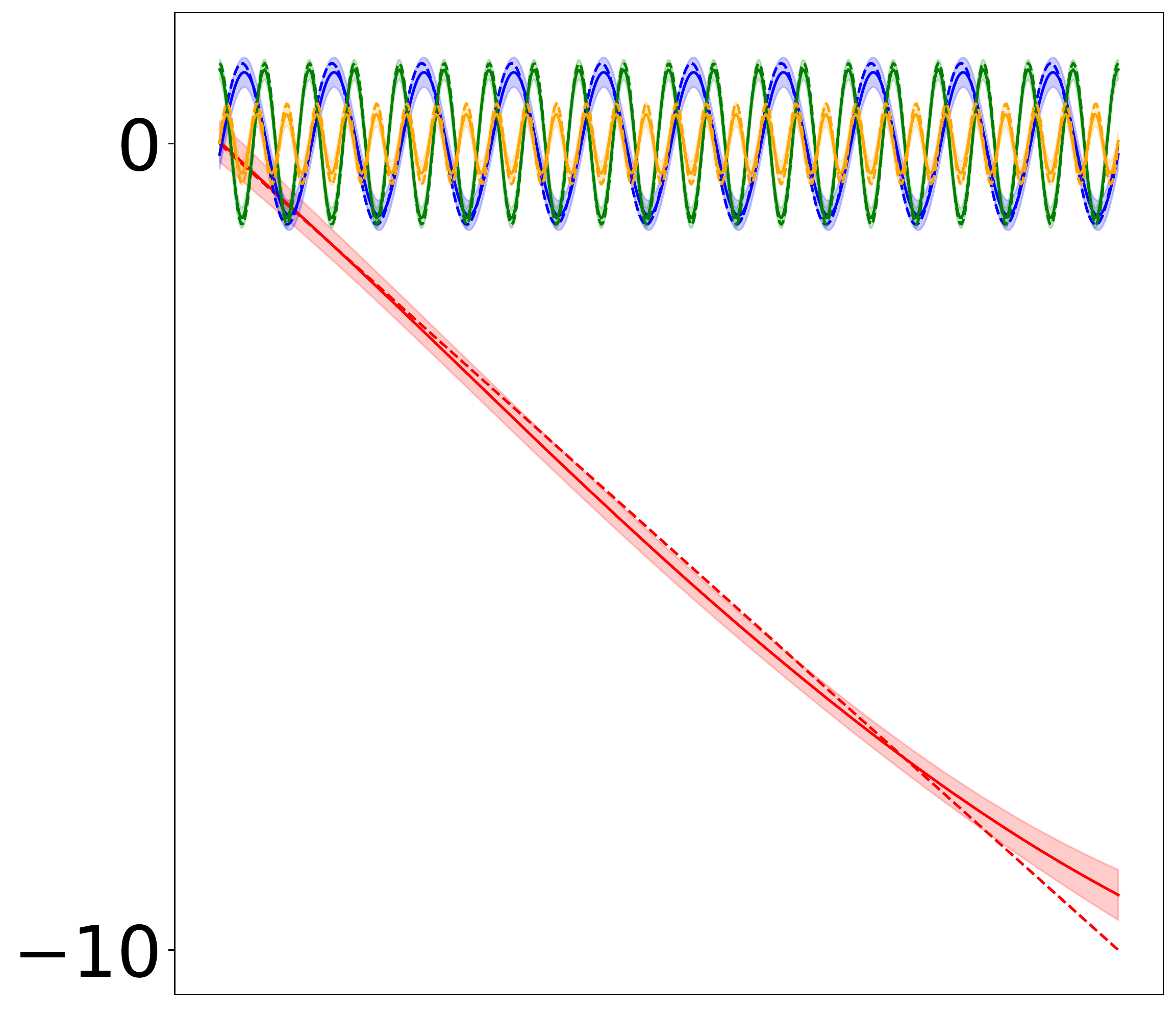}
            \caption{\small Channel\#4's factors}
            \label{fig:simu-channel-4}
		\end{subfigure} 
		\end{tabular}
		\vspace{0.03in}
		\caption{\small (a): The multivariate time series recovered from observations. The shaded region indicates two posterior standard deviations. (b)-(e): The weighted trend-seasonality factors learned by \ours of each channel.}
		\label{fig:core-simu}
	\end{figure*}

The imputation results are shown in Figure \ref{fig:simu-impute}. We can see that \ours recovers the series well, and the estimated uncertainty is reasonable. We also show the channel-wise estimated factors in Figure \ref{fig:simu-channel-1},\ref{fig:simu-channel-2},\ref{fig:simu-channel-3}, and \ref{fig:simu-channel-4}. We can see that the estimated factors are close to the real factors, which indicates that \ours can capture the underlying multiscale patterns of the data.

\subsection{Real-world Applications}

    \begin{table*}[]
        \centering
        \begin{small}
            \begin{tabular}{lccc|ccc|ccc}
                \toprule
                 {\textit{Observed-ratio}=50\%}& \multicolumn{3}{c}{\textit{Traffic-GuangZhou}} & \multicolumn{3}{c}{\textit{Solar-Power}} & \multicolumn{3}{c}{\textit{Uber-Move}} \\ 
                     {Metrics} & {RMSE} & {MAE} & {CRPS} & {RMSE} & {MAE} & {CRPS} & {RMSE} & {MAE} & {CRPS}\\
                    \midrule
                      \multicolumn{9}{l}{\textit{Deterministic \& Offline}}\\
                    \midrule
                    {SimpleMean} &{$9.852$} &{$7.791$} &{-} &{$3.213$}  &{$2.212$} &{-} &{$5.183$} &{$4.129$}  &{-} \\
                    {BRITS} &{$4.874$} &{$3.335$} &{-} &{$2.842$}  &{$1.985$} &{-} &{$2.180$} &{$1.527$} &{-}\\
                    {NAOMI} &{$5.986$} &{$4.543$} &{-} &{$2.918$}  &{$2.112$} &{-} &{$2.343$} &{$1.658$} &{-}\\
                    {SAITS} &{$4.839$} &{$3.391$} &{-} &{$2.791$}  &{$1.827$} &{-} &{$1.998$} &{$1.453$} &{-}\\
                    {TIDER} &{$4.708$} &{$3.469$} &{-} &\boldmath{$1.679$}  &{$0.838$} &{-} &{$1.959$} &{$1.422$} &{-}\\
                    \midrule
                      \multicolumn{9}{l}{\textit{Probabilistic \& Offline}}\\
                    \midrule
                    {Multi-Task GP} &{$4.887$} &{$3.530$} &{$0.092$} &{$2.847$}  &{$1.706$} &{$0.203$} &{$3.625$} &{$2.365$} &{$0.121$}\\
                    {GP-VAE}  &{$4.844$} &{$3.419$} &{$0.084$} &{$3.720$}  &{$1.810$} &{$0.368$} &{$5.399$} &{$3.622$} &{$0.203$}\\
                    {CSDI} &{$4.813$} &{$3.202$} &{$0.076$} &{$2.276$}  &{$0.804$} &{$0.166$} &{$1.982$} &{$1.437$} &{$0.072$} \\
                    {CSBI}  &{{$4.790$}} &{$3.182$} &{$0.074$} &{$2.097$}  &{$1.033$} &{$0.153$}  &{$1.985$} &{$1.441$} &{$0.075$} \\
                    \midrule
                      \multicolumn{9}{l}{\textit{Probabilistic \& Online}}\\
                    \midrule
                    {BayOTIDE-fix weight} &{{$11.032$}} &{$9.294$} &{$0.728$} &{$5.245$}  &{$2.153$} &{$0.374$}  &{$5.950$} &{$4.863$} &{$0.209$}  \\
                    {BayOTIDE-trend only} &{$4.188$} &{$2.875$} &{$0.059$} &{$1.789$}  &{$0.791$} &{$0.132$} &{$2.052$} &{$1.464$} &{$0.067$}\\
                    {BayOTIDE} &\boldmath{$3.820$} &\boldmath{$2.687$} &\boldmath{$0.055$} &{$1.699$}  &\boldmath{$0.734$} &\boldmath{$0.122$} &\boldmath{$1.901$} &\boldmath{$1.361$} &\boldmath{$0.062$} \\
                \bottomrule
            \end{tabular}
        \end{small}
        \caption{\small RMSE, MAE and CRPS scores of imputation results of all methods on three datasets with observed ratio $=50\%$. }
        \label{table:0.5-result}
    \end{table*}

\begin{table*}[]
    \centering
    \begin{small}
        \begin{tabular}{lccc|ccc|ccc}
            \toprule
             {\textit{Observed-ratio}=70\%}& \multicolumn{3}{c}{\textit{Traffic-GuangZhou}} & \multicolumn{3}{c}{\textit{Solar-Power}} & \multicolumn{3}{c}{\textit{UberLondon}} \\ 
                 {Metrics} & {RMSE} & {MAE} & {CRPS} & {RMSE} & {MAE} & {CRPS} & {RMSE} & {MAE} & {CRPS}\\
                \midrule
                  \multicolumn{9}{l}{\textit{Deterministic \& Offline}}\\
                \midrule
                {SimpleMean} &{$10.141$} &{$8.132$} &{-} &{$3.156$}  &{$2.319$} &{-} &{$5.323$} &{$4.256$}  &{-} \\
                {BRITS} &{$4.416$} &{$3.003$} &{-} &{$2.617$}  &{$1.861$} &{-} &{$2.154$} &{$1.488$} &{-}\\
                {NAOMI} &{$5.173$} &{$4.013$} &{-} &{$2.702$}  &{$2.003$} &{-} &{$2.139$} &{$1.423$} &{-}\\
                {SAITS} &{$4.407$} &{$3.025$} &{-} &{$2.359$}  &{$1.575$} &{-} &{$1.893$} &{$1.366$} &{-}\\
                {TIDER} &{$4.168$} &{$3.098$} &{-} &{$1.676$}  &{$0.874$} &{-} &{$1.867$} &{$1.354$} &{-}\\
                \midrule
                  \multicolumn{9}{l}{\textit{Probabilistic \& Offline}}\\
                \midrule
                {Multi-Task GP} &{$4.471$} &{$3.223$} &{$0.082$} &{$2.618$}  &{$1.418$} &{$0.189$} &{$3.159$} &{$2.126$} &{$0.108$}\\
                {GP-VAE}  &{$4.373$} &{$3.156$} &{$0.075$} &{$3.561$}  &{$1.723$} &{$0.331$} &{$3.133$} &{$2.005$} &{$0.625$}\\
                {CSDI} &{$4.301$} &{$2.991$} &{$0.069$} &{$2.132$}  &{$1.045$} &{$0.153$} &{$1.886$} &{$1.361$} &{$0.068$} \\
                {CSBI}  &{{$4.201$}} &{$2.955$} &{$0.064$} &{$1.987$}  &{$0.926$} &{$0.138$}  &{$1.899$} &{$1.353$} &{$0.070$} \\
                \midrule
                  \multicolumn{9}{l}{\textit{Probabilistic \& Online}}\\
                \midrule
                {BayOTIDE-fix weight} &{{$13.319$}} &{$9.29$} &{$0.677$} &{$5.238$}  &{$2.026$} &{$0.388$}  &{$5.889$} &{$4.849$} &{$0.208$}  \\
                {BayOTIDE-trend only} &{$4.002$} &{$2.759$} &{$0.056$} &{$1.651$}  &{$0.712$} &{$0.124$} &{$2.015$} &{$1.438$} &{$0.065$}\\
                {BayOTIDE} &\boldmath{$3.724$} &\boldmath{$2.611$} &\boldmath{$0.053$} &\boldmath{$1.621$}  &\boldmath{$0.709$} &\boldmath{$0.116$} &\boldmath{$1.832$} &\boldmath{$1.323$} &\boldmath{$0.061$} \\
            \bottomrule
        \end{tabular}
    \end{small}
    \caption{\small RMSE, MAE and CRPS scores of imputation results of all methods on three datasets with observed ratio $=70\%$.}
    \label{table:0.7-result}
\end{table*}

\textbf{Datasets} We evaluate \ours on three real-world datasets, \textit{Traffic-Guangzhou}\citep{chen10urban}: traffic speed records in Guangzhou with $214$ channels and $500$ timestamps. \textit{Solar-Power}(\url{https://www.nrel.gov/grid/solar-power-data.html}) : $137$ channels and $52560$ timestamps, which records the solar power generation of $137$ PV plants. \textit{Uber-Move}(\url{https://movement.uber.com/}): $7489$ channels and $744$ timestamps, recording the average movement of Uber cars along with the road segments in London, Jan 2020.  For each dataset, we randomly sample $\{70\%, 50\%\}$ of the available data points as observations for model training, and the rest for evaluation. The data process and split strategy are aligned with TIDER \citep{liu2023multivariate}.

\textbf{Baselines and Settings}
To the best of our knowledge, there are no online algorithms for multivariate time series imputation. Thus,  we set several popular deterministic and probabilistic offline imputation approaches as baselines. The deterministic group includes:
(1) \textit{SimpleMean} \citep{acuna2004treatment}, impute with column-wise mean values. (2) \textit{BRITS} \citep{cao2018brits}, the RNN-based model for imputation with time decay (3)  \textit{NAOMI}\citep{liu2019naomi}, a Bidirectional RNN model build with adversarial training (4) \textit{SAITS}\citep{du2023saits}, a transformer-based model which adopts the self-attention mechanism.  (5) \textit{TIDER}\citep{liu2023multivariate}. State of art deterministic imputation model based on disentangled temporal representations. 

The probabilistic group includes: (1) \textit{Multi-Task GP}\citep{Bonilla08multitask}, the classical multi-output Gaussian process model (2) \textit{GP-VAE}\citep{fortuin2020gp}, a deep generative model which combines Gaussian Processes(GP) and variational autoencoder(VAE) for imputation (3)  \textit{CSDI}\citep{tashiro2021csdi} Famous probabilistic approach which apply conditional diffusion model to capture the temporal dependency. (4)\textit{CSBI} Advanced diffusion method that models the imputation task as a conditional Schrödinger Bridge(SB)\citep{chen2023provably}.  We also set \textit{BayOTIDE-fix-wight}  by fixing all weight values as one and \textit{BayOTIDE-trend-only}, and only using trend factor, respectively for \ours. 

For most baselines, we use the released implementation provided by the authors. We partially use the results of deterministic methods reported in \textit{TIDER}\citep{liu2023multivariate}, as the setting is aligned. To avoid the out-of-memory problem of diffusion-based and deep-based baselines, we follow the preprocess of original papers, which split the whole sequence into small patches and subsample the channels for those methods.  

For \ours, we implemented it by Pytorch and finetuned the $D_s, D_r$, and the kernel hyperparameters to obtain optimal results. Detailed information on hyperparameter settings is provided at Table \ref{tab:hyperparameter} in the appendix. For the metrics, we use the mean absolute error (MAE) and the root mean squared error (RMSE) as the deterministic evaluation metrics for all methods. We adopt the continuous ranked probability score (CRPS) and the negative log-likelihood (NLLK), for \ours and all probabilistic baselines. We use 50 samples from the posterior to compute CRPS and NLLK. We repeat all the experiments $5$ times and report the average results. For the CEP step at each timestamp in \ours, we use the \textit{damping trick} ~\citep{minka2001expectation} in several inner epochs to avoid numerical instability. For the online imputation, we use the online update equations to obtain the imputation results and run RTS smoother at every evaluation timestamp.

\begin{figure*}[t]
        \centering
        \begin{small}
        \setlength{\tabcolsep}{0pt}
        \begin{tabular}[c]{ccc}
			\begin{subfigure}[b]{0.31\textwidth}
				\centering
				\includegraphics[width=\linewidth]{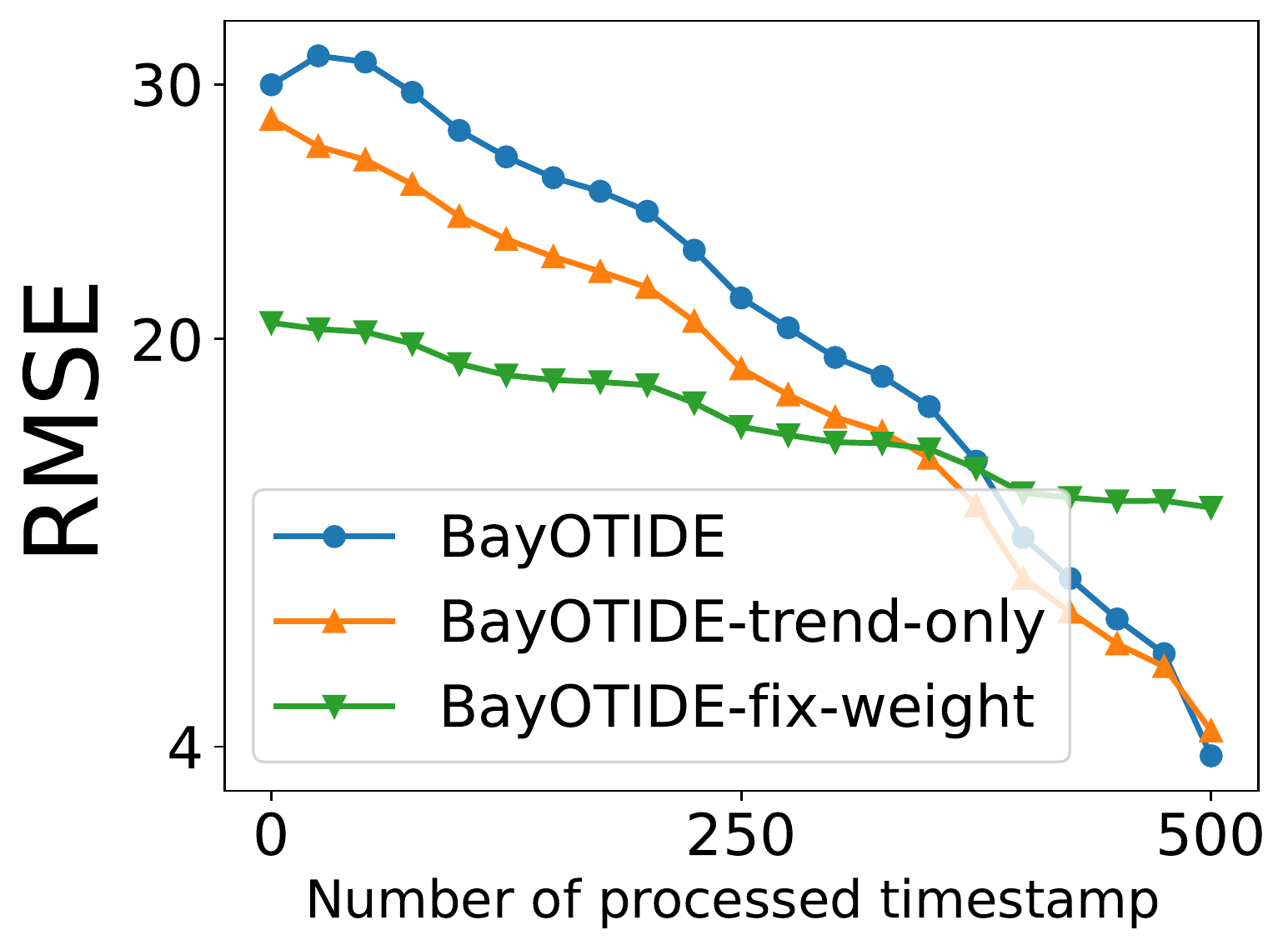}
                \caption{\small Online Imputation on Guangzhou}
				\label{fig:running_rmse-guangzhou}
            \end{subfigure} &
            \begin{subfigure}[b]{0.31\textwidth}
                \centering
                \includegraphics[width=\linewidth]{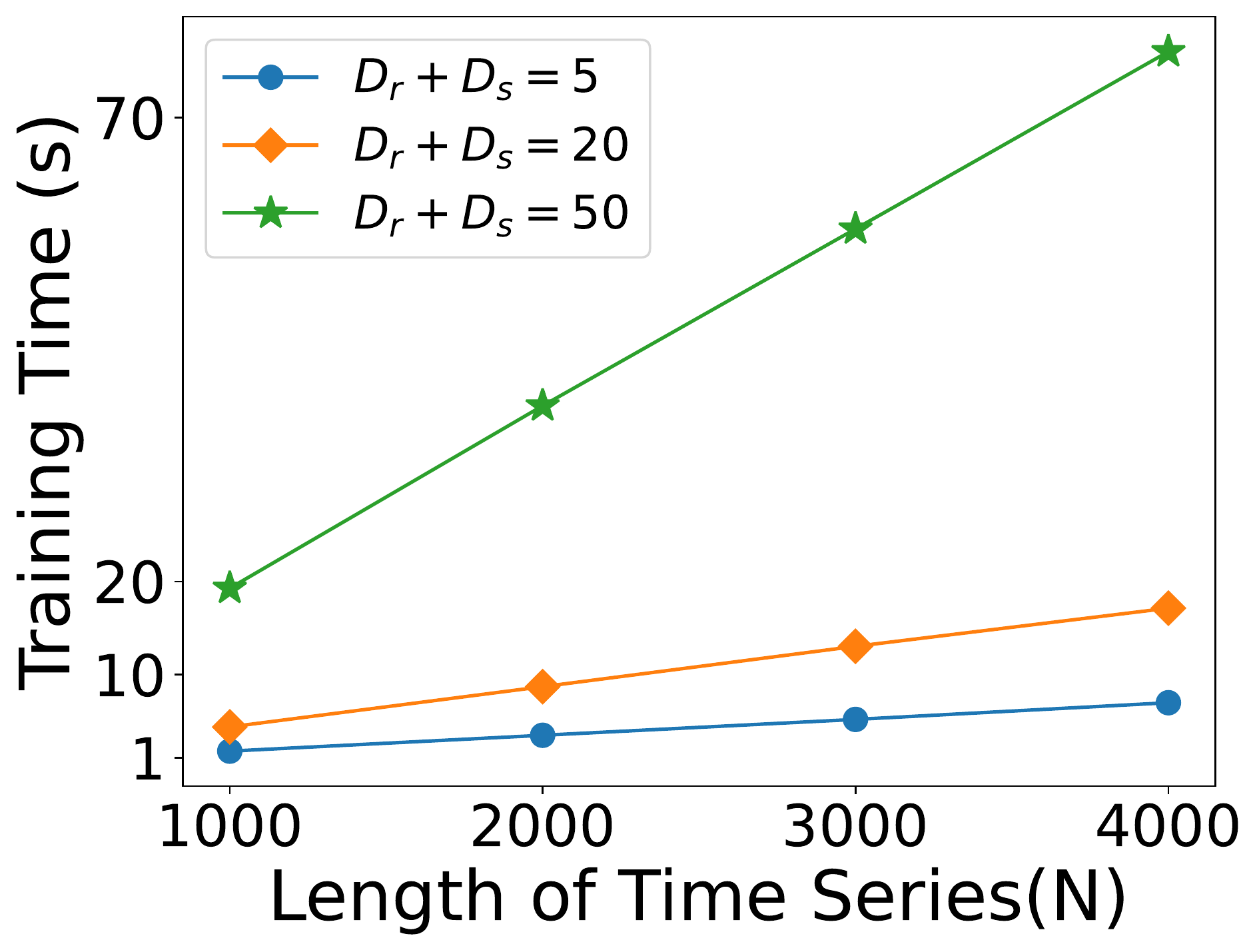}
                \caption{\small Scalability over time series length. }
                \label{fig:scalable-N}
            \end{subfigure} &
            \begin{subfigure}[b]{0.31\textwidth}
                \centering
                \includegraphics[width=\linewidth]{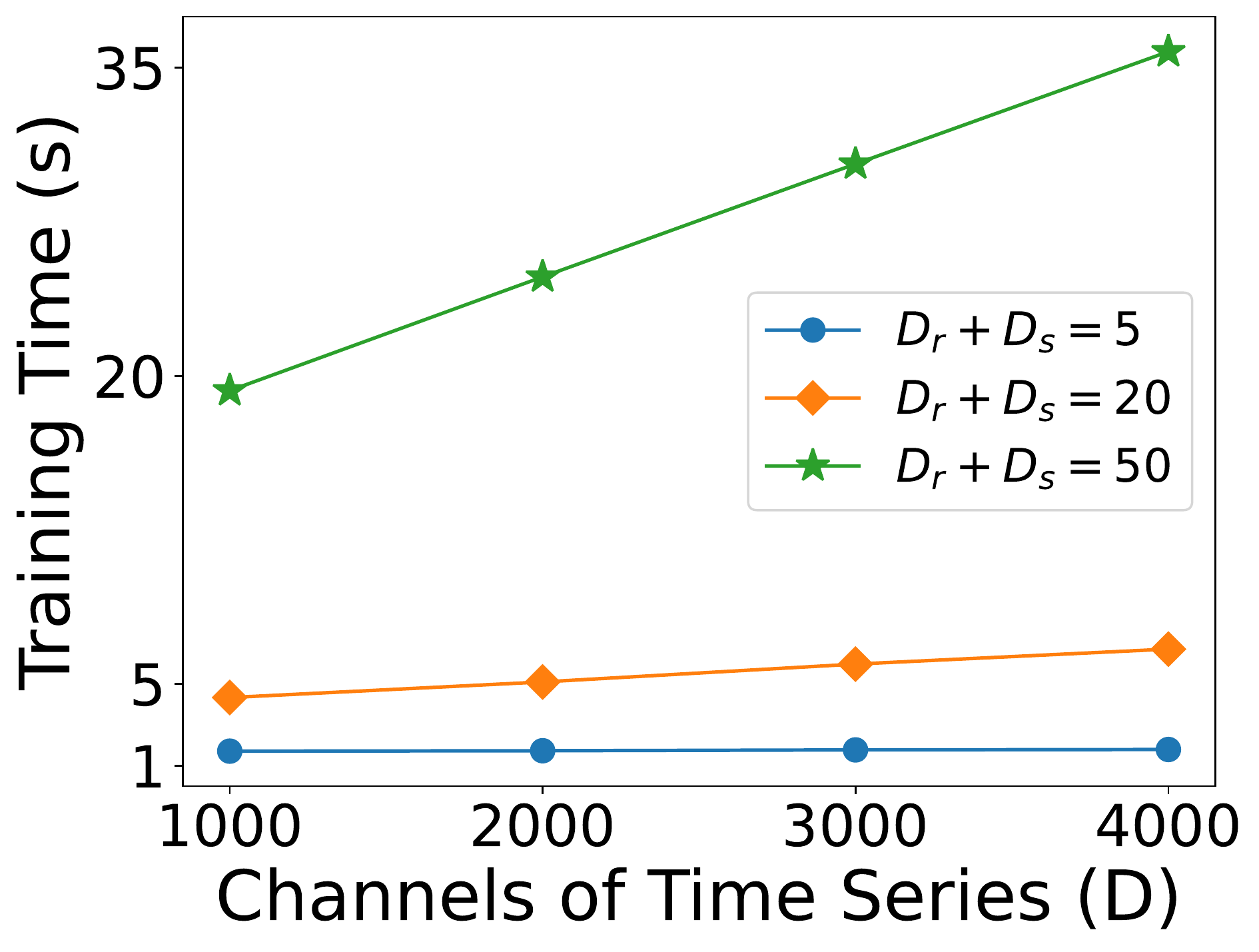}
                \caption{\small Scalability over the channels number.}
                \label{fig:scalable-D}
            \end{subfigure} \\
			\begin{subfigure}[b]{0.31\textwidth}
				\centering
				\includegraphics[width=\linewidth]{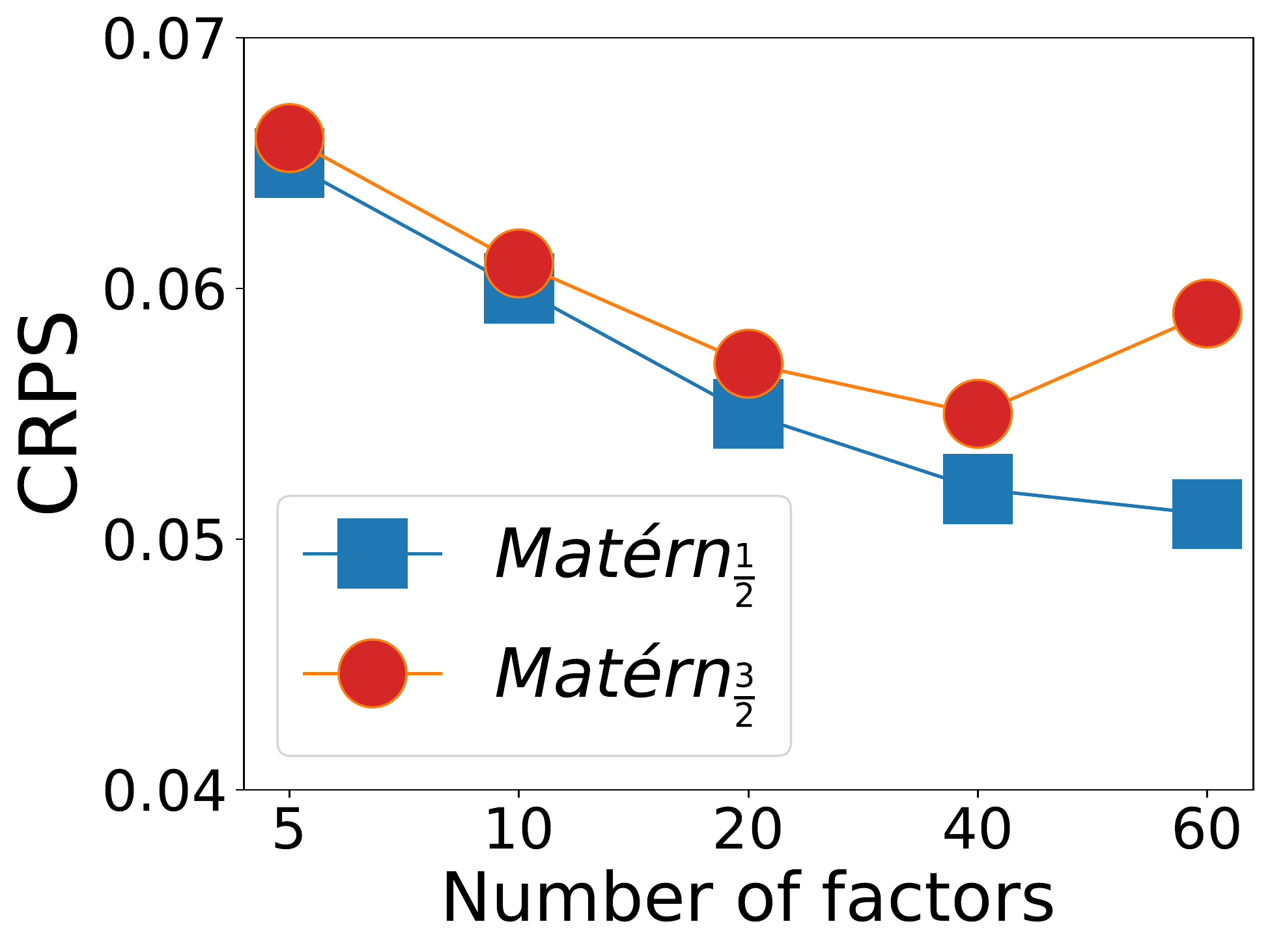}
                \caption{\small Sensitivity over factor number}
				\label{fig:sensitive-factor}
            \end{subfigure} &
            \begin{subfigure}[b]{0.31\textwidth}
                \centering
                \includegraphics[width=\linewidth]{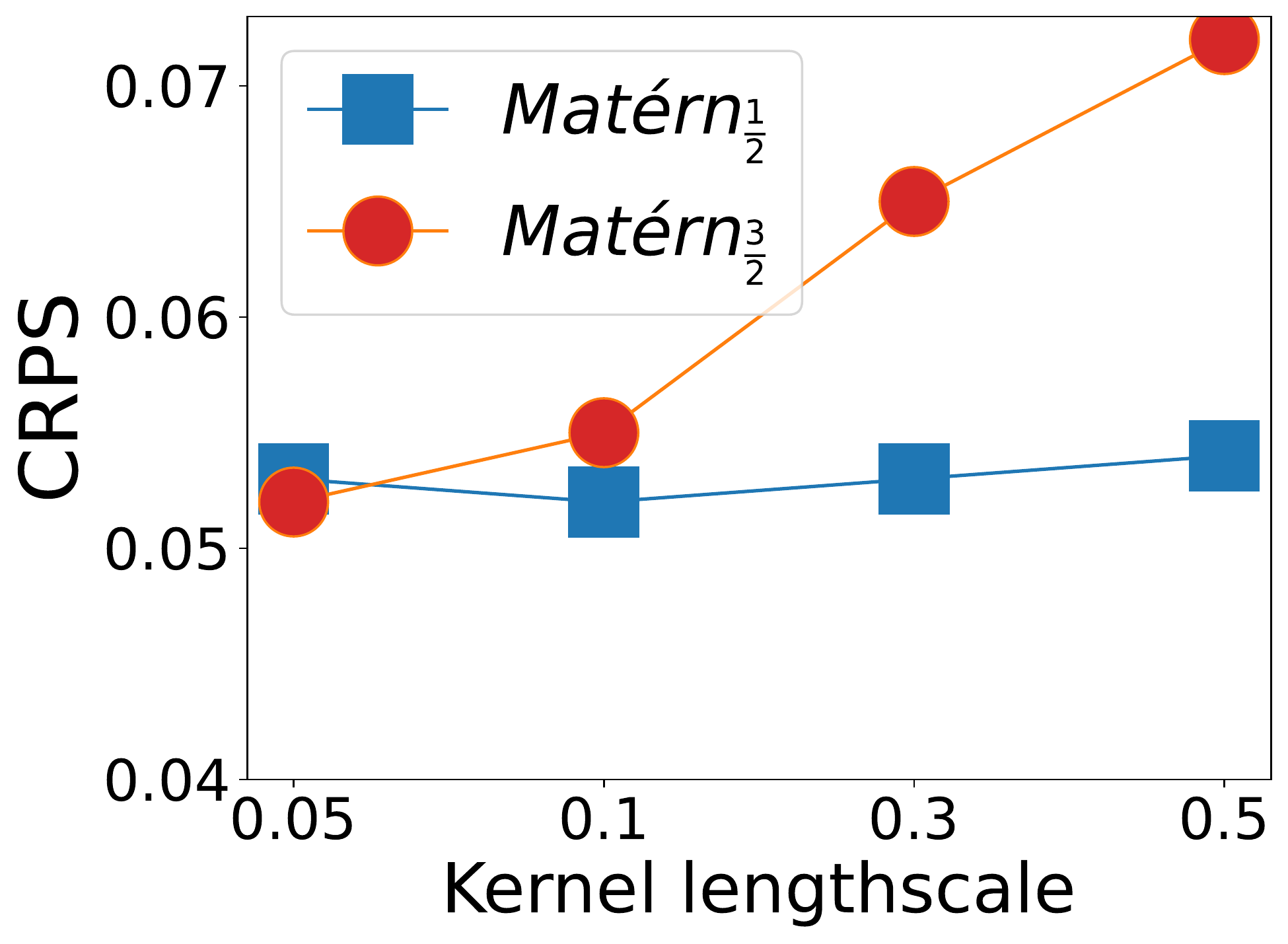}
                \caption{\small Sensitivity over kernel lengthscale}
                \label{fig:sensitive-lengthscale}
            \end{subfigure} &
            \begin{subfigure}[b]{0.31\textwidth}
                \centering
                \includegraphics[width=\linewidth]{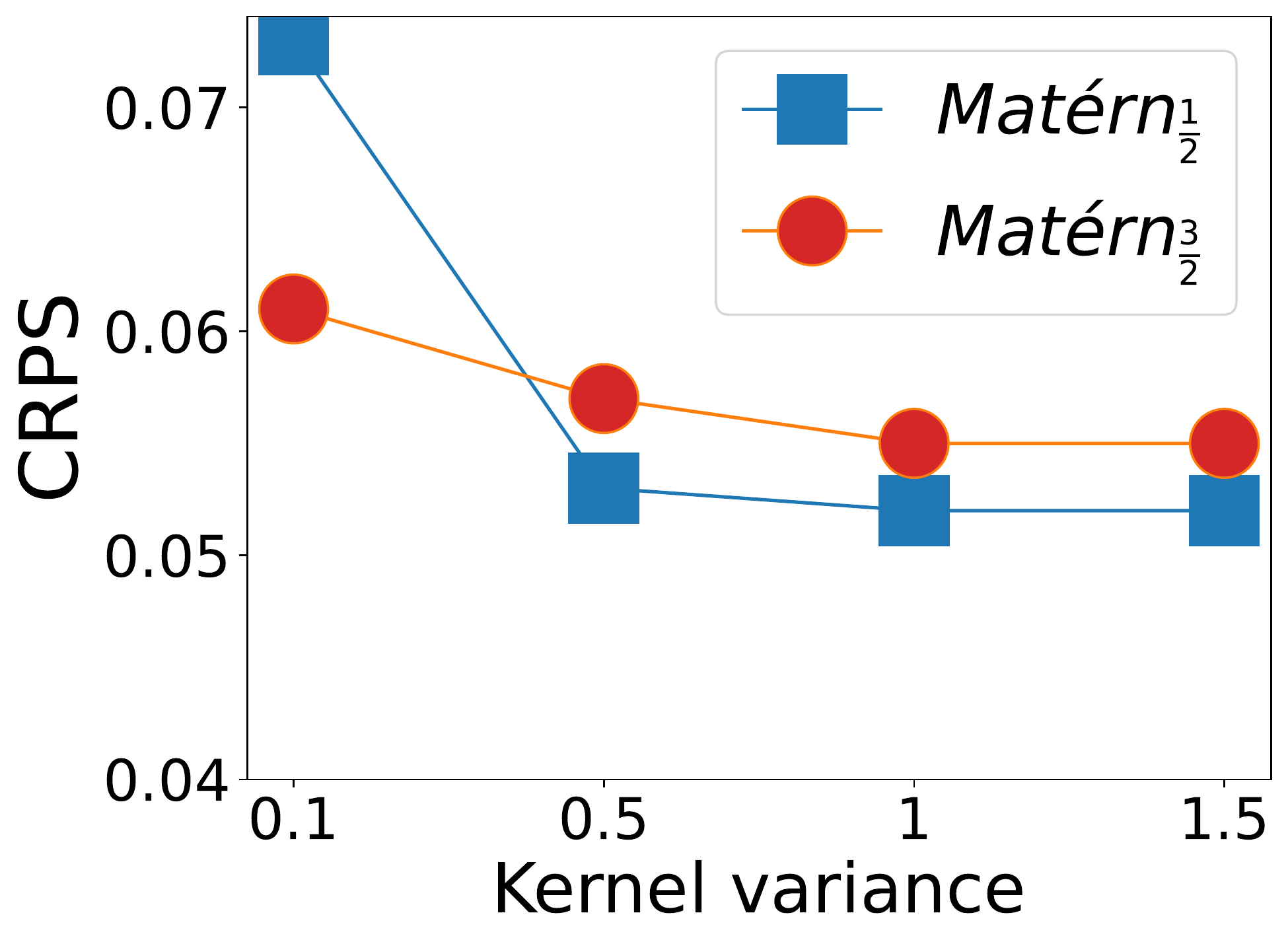}
                \caption{\small  Sensitivity over kernel variance}
                \label{fig:sensitive-variance}
            \end{subfigure} 
        \end{tabular} 
        \caption{ Online performance, scalability and sensitivity of \ours}
    \end{small}
\end{figure*}

\textbf{Deterministic and Probabilistic performance}
Table \ref{table:0.5-result} and table \ref{table:0.7-result} show the RMSE, MAE, and CRPS scores of imputation on three datasets with observed ratio $=50\%$ and $=70\%$ respectively. We can see that \ours, an online method that only processes data once, beats the offline baselines and performs best in most cases. \textit{TIDER} is the sub-optimal method for most cases. For probabilistic approaches, diffusion-based \textit{CSDI} and \textit{CSBI} obtain fair performance, but are costly in memory and need to split the long sequence into small patches for training. \textit{BayOTIDE-fix-wight} is with poor performance, which indicates that the weighted bases mechanism is crucial. \textit{BayOTIDE-trend-only} is slightly worse than \ours, showing the modeling of periodic factor is necessary. The results of  NLLK score over three datasets can be found at table \ref{table:nllk} in the appendix.

\textbf{Online Imputation}
We demonstrate the online imputation performance of \ours on three datasets with observed ratio $50\%$. Whenever a group of observations at new timestamps have been sequentially fed into the model, we evaluate the test RMSE of the model with the updated weights and temporal factor. We compare the performance of \ours with the \textit{BayOTIDE-fix-wight}. The online result on \textit{Traffic-Guangzhou} is shown in Figure \ref{fig:running_rmse-guangzhou}. We can see that \ours shows the reasonable results that the evaluation error drops gradually when more timestamps are processed, meaning the model can continuously learn and improve. The performance of \textit{BayOTIDE-fix-wight} is very poor. It indicates the trivial usage of the GP-SS model for multivariate time series imputation may not be feasible. The online results for the other two datasets can be found in the appendix.

\textbf{Scalability and Sensitivity }
We evaluate the scalability of \ours over data size and factor numbers. We set three different factor numbers: $D_r + D_s = \{5,20,50\}$. As for the scalability over series length $N$, We make synthetic data with channel number $D=1000$, increase the $N$ from $1000$ to $4000$, and measure the training time. The result is shown in Figure \ref{fig:scalable-N}. Similarly, we fix the series length $N=1000$, increase the series channel $D$ from $1000$ to $4000$, and then measure the training time. The result is shown in Figure \ref{fig:scalable-D}. As we can see, the running time of \ours grows linearly in both channel and length size, and the factor number determines the slope. Therefore, \ours enjoys the linear scalability to the data size, which is suitable for large-scale applications.

We further examine the sensitivity of \ours  with different hyperparameters. We build the model with Matérn kernel with different smoothness $\nu=\{1 / 2, 3 / 2\}$ on the \textit{Traffic-Guangzhou} with observed ratio $70\%$. We vary three crucial hyperparameters: the number of factors $D_r + D_s$, the kernel length scale, and the kernel variance, and check how imputation performance (CRPS) changes. The results are shown in Figure \ref{fig:sensitive-factor}, Figure \ref{fig:sensitive-lengthscale}, and  Figure \ref{fig:sensitive-variance}, respectively. We found more factors that will result in better performance in general for both kernel types. We can find that the performance is relatively stable over different hyperparameters for Matérn kernel with smoothness $\nu=\{1 / 2\}$. For the Matérn kernel with smoothness $\nu=\{3 / 2\}$, the performance is more sensitive, especially to the kernel lengthscale. The smaller kernel lengthscale will result in better performance. The kernel variance is also sensitive, but the performance is relatively stable when the variance is large enough. More results on the sensitivity of \ours over different combinations of latent space dimensions  $(D_r, D_s)$ are shown in Table \ref{table:dim-test} in the appendix. 

\textbf{Imputation on Irregular and All-Channel-Missing Timestamps}
We further show \ours can work well with irregulate timestamps with functional and continuous design, and therefore can handle the challenging case of all-channel-missing case. We select the observations at $\{50\%,70\% \}$ randomly sampled irregulate timestamps for training, and evaluate the model on the left all-channel-missing timestamps. Such timestamps are never seen in the training, and the model needs to do the interpolation. We highlight that most existing advanced imputation methods cannot handle this hard case well. It's because they are based on the regular-time-interval setting, which assumes there is at least one observation at every timestamp during the training. However, \ours can apply the Eq. \eqref{eq:predictive} and give probabilistic imputation at the arbitrary continuous timestamp. Thus, we only list the results of \ours on three datasets in Table \ref{table:irregu-result}. We can see the performance is closed or even better than the standard imputation setting shown in Table \ref{table:0.5-result}.     \\

\begin{table}[h]
	\centering
	\begin{small}
		\begin{tabular}{lcc|cc|cc}
			\toprule
                 {Dataset} & \multicolumn{2}{c}{GuangZhou} &  \multicolumn{2}{c}{Solar-Power} &  \multicolumn{2}{c}{Uber-Move} \\
                 {Obs.}& $50 \%$ &  $70 \%$ & $50 \%$ &  $70 \%$ & $50 \%$ &  $70 \%$ \\ 
                \midrule
                {RMSE} &{$3.625$} &{$3.383$} &{$1.624$} &{$1.442$}  &{$3.017$} &{$2.931$} \\
                {MAE} &{$2.524$} &{$2.401$} &{$0.706$} &{$0.614$}  &{$1.199$} &{$1.154$} \\
                {CRPS} &{$0.051$} &{$0.046$} &{$0.121$} &{$0.113$}  &{$0.311$} &{$0.302$} \\
                {NLLK} &{$2.708$} &{$2.634$} &{$1.861$} &{$1.857$}  &{$2.137$} &{$2.138$} \\
			\bottomrule
		\end{tabular}
	\end{small}
	\caption{\small The imputation results of \ours with settings of irregulate and all-channel-missing timestamps on three datasets with observed ratio $=\{50\%,70\%\}$.}
	\label{table:irregu-result}
	\vspace{-0.1in}
\end{table}

\section{Conclusion}
We proposed \ours, a novel Bayesian model for online multivariate time series imputations. We decompose the multivariate time series into a temporal function basis and channel-wise weights, and apply a group of GPs to fit the temporal function basis. An efficient online inference algorithm is developed based on the SDE representation of GPs and moment-matching techniques. Results on both synthetic and real-world datasets show that \ours outperforms the state-of-the-art methods in terms of both imputation accuracy and uncertainty quantification. We also evaluated the scalability and robustness of \ours on large-scale real-world datasets with different settings of hyperparameters and missing patterns.
. In the future, we plan to extend \ours to handle more complex patterns and more challenging tasks, such as long-term forecasting with non-stationary patterns.

\newpage
\section*{Impact Statement}
This paper is dedicated to innovating time series imputation techniques to push the boundaries of time series analysis further. While our primary objective is to enhance imputation accuracy and computational efficiency, we are also mindful of the broader ethical considerations that accompany technological progress in this area. While immediate societal impacts may not be apparent, we acknowledge the importance of ongoing vigilance regarding the ethical use of these advancements. It is essential to continuously assess and address potential implications to ensure responsible development and application in various scenarios.

\bibliography{BayTIDE}
\bibliographystyle{icml2024}

\newpage
\newpage
\appendix
\section*{Appendix}

\subsection{LTI-SDE representation of GP with Matérn kernel and periodic kernel}
\subsubsection{Connect GP with LTI-SDE by Spectral Analysis} 
We take the Matérn kernel as an example to show how to connect GP with LTI-SDE. The Matérn kernel is defined as:

\begin{align} 
\kappa_\nu(t, t') = a \frac{(\frac{\sqrt{2 \nu}}{\rho} \Delta)^{\nu}}{\Gamma(\nu) 2^{\nu-1}} K_{\nu}(\frac{\sqrt{2 \nu}}{\rho} \Delta)
\end{align}

where $\Delta = |t - t'|$, $\Gamma(\cdot)$ is the Gamma function, $a>0$ and $\rho>0$ are the amplitude and length-scale parameters respectively, $K_{\nu}$ is the modified Bessel function of the second kind, and $\nu>0$ controls the smoothness of sample paths from the GP prior $f(t) \sim \mathcal{GP}(0, \kappa_\nu(t,t'))$. 

For a stationary Matérn kernel $\kappa_\nu(t, t') = \kappa_\nu(t - t')$, the energy spectrum density of $f(t)$ can be obtained via the Wiener-Khinchin theorem by taking the Fourier transform of $\kappa_\nu(\Delta)$:

\begin{align}
S(\omega) = \frac{\sigma^2}{(\alpha^2 + \omega^2)^{m+1}} 
\end{align}

where $\omega$ is the frequency, $\alpha = \frac{\sqrt{2\nu}}{\rho}$, and we take $\nu = m + \frac{1}{2}$ for $m \in \{0, 1, 2,...\}$. 

Expanding the polynomial gives:

\begin{align}
(\alpha + i \omega)^{m+1} = \sum_{k=0}^{m} c_k (i\omega)^k + (i\omega)^{m+1}
\end{align}

where $c_k$ are coefficients. This allows constructing an equivalent frequency domain system: 

\begin{align}
\sum_{k=1}^{m} c_k (i\omega)^k \widehat{f}(\omega) + (i\omega)^{m+1}\widehat{f}(\omega) = \widehat{\beta}(\omega)
\end{align}

where $\widehat{f}(\omega)$ and $\widehat{\beta}(\omega)$ are Fourier transforms of $f(t)$ and white noise ${w}(t)$ with spectral density $ q_\mathbf{s}$ respectively.

Taking the inverse Fourier transform yields the time domain SDE:

\begin{align}
\sum_{k=1}^{m} c_k \frac{d^k f}{dt^k} + \frac{d^{m+1}f}{dt^{m+1}} = {w}(t) \label{eq:high-order-sde}
\end{align}
 We can further construct a new state $\z = (f, f^{(1)}, \ldots, f^{(m)})^\top$ (where each $f^{(k)} \overset{\Delta}{=} \d^k f/\d t^k$) and convert \eqref{eq:high-order-sde} into a linear time-invariant (LTI) SDE, 
\begin{align}
	\frac{\mathrm{d} \mathbf{z}(t)}{\mathrm{d} t}= \mathbf{F} \mathbf{z}(t) + \mathbf{L} {w}(t) \label{eq:LTI-SDE-2}
\end{align}
where 
\begin{align}
	\F =\left(\begin{array}{cccc}
		0 & 1 & & \\
		& \ddots  & \ddots & \\
		&   & 0 & 1 \\
		-c_0 & \ldots & -c_{m-1} & -c_{m}
	\end{array}\right), \quad \L =\left(\begin{array}{c}
		0 \\
		\vdots\\
		0 \\
		1\\
	\end{array}\right). \notag 
\end{align}

The LTI-SDE is particularly useful in that its finite set of states follows a Gauss-Markov chain, namely the state-space prior.  Specifically, given arbitrary $t_1 < \ldots < t_L$, we have 
\begin{align}
	p(\z(t_1), \ldots, \z(t_L)) = p(\z(t_1)) \prod\nolimits_{k=1}^{L-1} p(\z(t_{k+1})| \z(t_{k})) \notag 
\end{align}
where
\begin{align}
	p(\z(t_1)) &= \N(\z(t_1)|\0, \P_\infty), \notag \\
	p(\mathbf{z}(t_{n+1}) | \mathbf{z}(t_{n})) &= \mathcal{N}(\mathbf{z}(t_{n+1}) | \mathbf{A}_n \mathbf{z}(t_{n}),\mathbf{Q}_n ) \label{eq:SS-prior}
\end{align}
where $\mathbf{A}_n =  \operatorname{exp}(\mathbf{F}\Delta_n)$, $\mathbf{Q}_n = \int_{t_n}^{t_{n+1}} \mathbf{A}_n \mathbf{L} \mathbf{L}^{\top} \mathbf{A}_n^{\top} q_\mathbf{s} \mathrm{d} t$, $\Delta_n = t_{n+1} - t_n$, and $\mathbf{P_{\infty}}$ is the steady-state covariance matrix of the LTI-SDE \ref{eq:lti-sde}, which can be obtained by solving the Lyapunov equation $\mathbf{F} \mathbf{P_{\infty}} + \mathbf{P_{\infty}} \mathbf{F}^{\top} + \mathbf{L} \mathbf{L}^{\top} q_\mathbf{s}= 0$ \citep{lancaster1995algebraic}, as we claimed in the main paper.

Note that for other types of stationary kernel functions, such as the periodic kernels, we can approximate the inverse spectral density $1/S(\omega)$ with a polynomial of $\omega^2$ with negative roots \citep{solin2014explicit}, and follow the same way to construct an LTI-SDE and state-space prior. 

\subsubsection{The closed-form of LTI-SDE and state space prior with Matérn kernel and periodic kernel}

With the canonical form of LTI-SDE \eqref{eq:LTI-SDE-2}and state space prior\eqref{eq:SS-prior} and above derivation, we can work out the closed-form of LTI-SDE and state space prior for Matérn kernel and periodic kernel. We present the results in the following.

For Matérn kernel with $m=0$, indicating the smoothness is $\nu = 0+\frac{1}{2}$, it becomes the exponential covariance function:
\begin{align}
    \kappa_{\exp }(\tau)=\sigma^2 \exp \left(-\frac{\tau}{\ell}\right)
\end{align}
Then the parameters of the LTI-SDE and state space prior are:  $\{m=0, \mathbf{F}= -1/l, \mathbf{L}=1, q_\mathbf{s}=2\sigma^2/l,\mathbf{P_{\infty}}=\sigma^2\}$

For Matérn kernel with $m=1$, indicating the smoothness is $\nu = 1+\frac{1}{2}= 3/2$, the kernel becomes the Matérn 3/2 covariance function:   
\begin{align}
    \kappa_{\text {Mat. }}(\tau)=\sigma^2\left(1+\frac{\sqrt{3} \tau}{\ell}\right) \exp \left(-\frac{\sqrt{3} \tau}{\ell}\right)
\end{align}
and the parameters of the LTI-SDE and state space prior are: $m=1$, $\mathbf{F}=\left(\begin{array}{cc}
    0 & 1 \\
    -\lambda^2 & -2 \lambda
    \end{array}\right)$,$
    \mathbf{L}=\left(\begin{array}{l}
    0 \\
    1
    \end{array}\right)
    $,
    $\mathbf{P}_{\infty}=\left(\begin{array}{cc}\sigma^2 & 0 \\ 0 & \lambda^2 \sigma^2\end{array}\right)$, $q_\mathbf{s} = 4 \lambda^3 \sigma^2$, where $\lambda=\sqrt{3} / \ell$.

    For the periodic kernel:
\begin{align}
    \kappa_{\text{periodic}}(t, t') = \sigma^2 \exp\left(-\frac{2\sin^2(\pi\Delta/p)}{l^2}\right)\label{eq:periodic-kernel-2}
\end{align}
with preset periodicity $p$, \citep{solin2014explicit} construct corresponding SDE by a sum of n two-dimensional SDE models(m=1) of the following parameters:
\begin{align}
    \mathbf{F}_j=\left(\begin{array}{cc}
        0 & -\frac{2\pi}{p} j \\
        \frac{2\pi}{p} j & 0
        \end{array}\right), \mathbf{L}_j=\left(\begin{array}{cc}
            1 & 0 \\
            0 & 1
            \end{array}\right)
\end{align}
$\mathbf{P}_{\infty, j}=q_j^2 \mathbf{I}_2$, where $q_j^2=2 \mathrm{I}_j\left(\ell^{-2}\right) / \exp \left(\ell^{-2}\right)$, for $j=1,2, \ldots, n$ and $q_0^2=\mathrm{I}_0\left(\ell^{-2}\right) / \exp \left(\ell^{-2}\right)$ ~\citep{solin2016stochastic}

\subsection{Derivative of online update equations by conditional moment matching}
\subsubsection{Brief Introduction of EP and CEP }

The Expectation Propagation (EP) ~\citep{minka2001family} and Conditional EP (CEP) ~\citep{wang2019conditional} frameworks approximate complex probabilistic models with distributions in the exponential family.

Consider a model with latent variables $\btheta$ and observed data $\Dcal = \{\y_1, \ldots, \y_N\}$. The joint probability is:  

\begin{align}
p(\btheta, \Dcal) = p(\btheta) \prod_{n=1}^N p(\y_n|\btheta) 
\end{align}

The posterior $p(\btheta|\Dcal)$ is usually intractable. EP approximates each term with an exponential family distribution:

\begin{align}
p(y_n|\btheta) &\approx c_n f_n(\btheta) \\  
p(\btheta) &\approx c_0 f_0(\btheta)
\end{align}

where $f_n(\btheta) \propto \exp(\blambda_n^\top \bphi(\btheta))$ are in the exponential family with natural parameters $\blambda_n$ and sufficient statistics $\bphi(\btheta)$.

The joint probability is approximated by: 

\begin{align}
p(\btheta, \Dcal) \approx f_0(\btheta) \prod_{n=1}^N f_n(\btheta) \cdot \text{const}
\end{align}

giving a tractable approximate posterior $q(\btheta) \approx p(\btheta|\Dcal)$. 

EP optimizes the approximations $f_n$ by repeatedly:

1) Computing the calibrated distribution $q^{\backslash n}$ excluding $f_n$.

2) Constructing the tilted distribution $\tp$ incorporating the true likelihood.

3) Projecting $\tp$ back to the exponential family by moment matching. 

4) Updating $f_n \approx \frac{q^{*}}{q^{\backslash n}}$ where $q^{*}$ is the projection.

The intractable moment matching in Step 3 is key. CEP exploits factorized $f_n = \prod_m f_{nm}(\btheta_m)$ with disjoint $\btheta_m$. It uses nested expectations:

\begin{align}
\EE_\tp[\bphi(\btheta_m)] = \EE_{\tp(\btheta_{\backslash m})} \EE_{\tp(\btheta_m|\btheta_{\backslash m})} [\bphi(\btheta_m)]
\end{align} 

The inner expectation is tractable. For the outer expectation, CEP approximates the marginal tilted distribution with the current posterior: 

\begin{align}
\EE_{\tp(\btheta_{\backslash m})} [\g(\btheta_{\backslash m})] \approx \EE_{q(\btheta_{\backslash m})}[\g(\btheta_{\backslash m})]
\end{align}

If still intractable, the delta method is used to approximate the expectation with a Taylor expansion.

Once the conditional moment $\g(\btheta_{\backslash m})$ is obtained, CEP substitutes the expectation $\EE_{q(\btheta_{\backslash m})}[\btheta_{\backslash m}]$ to compute the matched moment for constructing $q^*$.

\subsection*{Online inference Update}
We then applied the EP and CEP  to approximate the running posterior $p(\Theta \mid \mathcal{D}_{t_n} \cup \mathbf{y}_{n+1}) $. With the incremental version of Bayes'rule \eqref{eq:online-inference}, the key is to work out the close-form factors in the likelihood approximation  \eqref{eq:approx-likelihood}.  In other words, we adopt conditional moment match techniques to handle:
\begin{align}
    &\mathcal{N}(\mathbf{y}_{n+1}^{d} \mid \mathbf{u}^{d} \mathbf{V}(t_{n+1})  , \tau^{-1})  \approx \nonumber\\ 
    & \mathcal{Z}  f_{n+1}^{d}(\mathbf{Z}(t_{n+1}))f_{n+1}^{d}(\u_d)f_{n+1}^{d}(\tau) 
\end{align}
Then we follow the standard CEP procedure to compute the conditional moment of $\{\mathbf{Z}(t_{n+1}), \u_d, \tau \}$ and update  $f_{n+1}^{d}(\u_d) = \mathcal{N}(\u_d \mid \mathbf{\hat{m}}_{n+1}^{d}, \mathbf{\hat{V}}_{n+1}^{d})$ and $f_{n+1}^{d}(\tau) = \operatorname{Gamma}(\tau \mid \hat{a}_{n+1},\hat{ b}_{n+1})$,  $f_{n+1}^{d}(\mathbf{Z}(t_{n+1})) = \operatorname{concat}[\mathcal{N}(\hat{\mu}_i, \hat{\mathbf{S}_i})]$. 

Specifically, for $f_{n+1}^{d}(\tau) = \operatorname{Gamma}(\tau \mid \hat{a}_{n+1},\hat{ b}_{n+1})$ we have: 
\begin{align}
   & \hat{a}_{n+1} = \frac{1}{2} \\
   &  \hat{b}_{n+1} = \frac{1}{2} \mathbb{E}_{q}[ (\mathbf{y}_{n+1}^{d} - \mathbf{u}^{d} \mathbf{V}(t_{n+1}))^2  ]
\end{align}

For $f_{n+1}^{d}(\u_d) = \mathcal{N}(\u_d \mid \mathbf{\hat{m}}_{n+1}^{d}, \mathbf{\hat{V}}_{n+1}^{d})$, we have:
\begin{align}
   & \mathbf{\hat{V}}_{n+1}^{d} = [  \mathbb{E}_{q}[\tau \cdot  \mathbf{Z}(t_{n+1}) \mathbf{Z}^{T}(t_{n+1})]  ]^{-1}\\
   & \mathbf{\hat{m}}_{n+1}^{d} =  \mathbf{\hat{V}}_{n+1}^{d} \cdot  \mathbb{E}_{q}[ \tau \mathbf{y}_{n+1}^{d}  \mathbf{Z}(t_{n+1}) ] 
\end{align}

For  $f_{n+1}^{d}(\mathbf{Z}(t_{n+1})) = \operatorname{concat}[\mathcal{N}(\hat{\mu}_i, \hat{{S}_i})] = \mathcal{N}(\mathbf{\hat{\mu}_i}, \hat{\mathbf{S}_i})$ , we have:
\begin{align}
    &\hat{\mathbf{S}_i} = [  \mathbb{E}_{q}[\tau \cdot \u_d\u_d^{T}]]^{-1}\\
    &\mathbf{\hat{\mu}_i} = \hat{\mathbf{S}_i}  \cdot  \mathbb{E}_{q}[ \tau \mathbf{y}_{n+1}^{d} \u_d ] 
\end{align}

All the expectation is taken over the current approximated posterior $q(\Theta \mid \mathcal{D}_{t_n} ) $. 

With these message factors from the new-arriving likelihood, the online update is easy. We follow the \eqref{eq:update-tau},  \eqref{eq:update-U} and \eqref{eq:update-Z} to merge the factors and obtain the closed-form online update for the global posterior.

\subsection{ Derivation of the PROBABILISTIC IMPUTATION AT ARBITRARY TIME STAMPS}
Consider a general state space model, which includes a sequence of states $\x_1, \ldots, \x_M$ and the observed data $\Dcal$. The states are at time $t_1, \ldots, t_M$ respectively. 
The key of the state space model is that the prior of the states is a Markov chain. The joint probability has the following form, 
\begin{align}
    & p(\x_1, \ldots, \x_M, \Dcal) \nonumber \\
    & = p(\x_1)\prod_{j=1}^{M-1} p(\x_{j+1}|\x_j) \cdot p(\Dcal|\x_1, \ldots, \x_M). \label{eq:ss1}
\end{align}
Note that here we do not assume the data likelihood is factorized over each state, like those typically used in Kalman filtering. In our point process model, the likelihood often couples multiple states together. 

Suppose we have run some posterior inference to obtain the posterior of these states $q(\x_1, \ldots, \x_M)$, and we can easily pick up the marginal posterior of each state and each pair of the states. Now we want to calculate the posterior distribution of the state at time $t^*$ such that $t_m < t^* < t_{m+1}$. Denote the corresponding state by $\x^*$, our goal is to compute $p(\x^*|\Dcal)$. To do so, we consider incorporating $\x^*$ in the joint probability \eqref{eq:ss1}, 
\begin{align}
    &p(\x_1, \ldots, \x_{m}, \x^*, \x_{m+1}, \ldots, \x_M, \Dcal) \notag \\
    &= p(\x_1)\prod_{j=1}^{m-1} p(\x_{j+1}|\x_j) \cdot p(\x^*|\x_m) p(\x_{m+1}|\x^*) \notag \\
    &\cdot  \prod_{j={m+1}}^Mp(\x_{j+1}|\x_j)\cdot p(\Dcal|\x_1, \ldots, \x_M). \label{eq:ss2}
\end{align}
Now, we marginalize out $\x_{1:M \setminus \{m, m+1\}} = \{\x_1, \ldots, \x_{m-1}, \x_{m+2}, \ldots, \x_{M}\}$. Note that since $\x^*$ does not appear in the likelihood, we can take it out from the integral, 
\begin{align}
     &p(\x_m, \x_{m+1}, \x^*, \Dcal)  \notag \\
     &=\int p(\x_1)\prod_{j=1}^{m-1} p(\x_{j+1}|\x_j)\notag \\
     &  \prod_{j={m+1}}^Mp(\x_{j+1}|\x_j)\cdot p(\Dcal|\x_1, \ldots, \x_M) \d \x_{1:M\setminus\{m, m+1\}} \notag \\
     &\cdot p(\x^*|\x_m) p(\x_{m+1}|\x^*) \notag \\
     &=\frac{p(\x_m, \x_{m+1}, \Dcal)p(\x^*|\x_m)p(\x_{m+1}|\x^*)}{p(\x_{m+1}|\x_m)}.
\end{align}

Therefore, we have 
\begin{align}
    & p(\x_m, \x_{m+1}, \x^*|\Dcal) \propto \nonumber \\ & p(\x_m, \x_{m+1}|\Dcal) p(\x^*|\x_m)p(\x_{m+1}|\x^*).
\end{align}
Suppose we are able to obtain $p(\x_m, \x_{m+1}|\Dcal) \approx q(\x_m, \x_{m+1})$. We now need to obtain the posterior of $\x^*$. In the LTI SDE model, we know that the state transition is a Gaussian jump. Let us denote
\begin{align}
    &p(\x^*|\x_m) = \N(\x^* | \A_1 \x_m, \Q_1) \nonumber \\
    &p(\x_{m+1}|\x^*) = \N(\x_{m+1} | \A_2 \x^*, \Q_2).
\end{align}
Then, we can simply merge the natural parameters of the two Gaussian and obtain
\begin{align}
p(\x_m, \x_{m+1}, \x^*|\Dcal) = p(\x_m, \x_{m+1}|\Dcal) \N(\x^*|\bmu^*, \bSigma^*), \label{eq:post3} 
\end{align}
where 
\begin{align}
    \left(\bSigma^*\right)^{-1} &= \Q_1^{-1} + \A_2^\top \Q_2^{-1} \A_2, \notag \\
    \left(\bSigma^*\right)^{-1}\bmu^* &= \Q_1^{-1} \A_1 \x_m + \A_2^\top \Q_2^{-1}\x_{m+1}. \label{eq:cond-moment}
\end{align}
Those are the formulas for probabilistic imputation as arbitrary time stamps.

\begin{table*}[]
    \centering
    \begin{small}
        \begin{tabular}{lccc}
            \hline \textit{Observed-ratio}=50\% & \textit{Traffic-GuangZhou} & \textit{Solar-Power} & \textit{Uber-Move} \\
            \hline 
            Number of trend factor $D_r$ & $30$ & $50$ & $30$ \\
            Number of seasonality factor $D_s$ & $10$ & $5$ & $5$ \\
            Trend factor smoothness ($\nu$) & $\frac{1}{2}$ & $\frac{3}{2}$ & $\frac{3}{2}$ \\
            Trend factor lengthscale & $0.1$ &  $0.001$ &  $0.1$ \\
            Trend factor variance & $1.0$ & $1.0$ & $1.0$ \\
            Seasonal factor frequency ($2\pi p$)& $15$ & $10$ & $15$\\
            Seasonal factor lengthscale & $0.05$ & $0.5$ & $0.05$ \\
            Damping epochs & $5$ & $2$ &$5$ \\

            \hline \hline \textit{Observed-ratio}=70\% & \textit{Traffic-GuangZhou} & \textit{Solar-Power} & \textit{Uber-Move} \\
            \hline 
            Number of trend factor $D_r$ & $30$ & $50$ & $30$ \\
            Number of seasonality factor $D_s$ & $10$ & $5$ & $5$ \\
            Trend factor smoothness ($\nu$) & $\frac{1}{2}$ & $\frac{1}{2}$ & $\frac{1}{2}$ \\
            Trend factor lengthscale & $0.1$ &  $0.0005$ &  $0.1$ \\
            Trend factor variance & $1.0$ & $1.0$ & $1.0$ \\
            Seasonal factor frequency ($2\pi p$)& $15$ & $100$ & $15$\\
            Seasonal factor lengthscale & $0.05$ & $0.5$ & $0.05$ \\
            Damping epochs & $5$ & $2$ &$5$ \\
            \hline
            \end{tabular}
\end{small}
\caption{\small The hyperparameter setting of \ours for the imputation task.}
\label{tab:hyperparameter}
\end{table*}

\subsection{More experimental results}

The NLLK scores of probabilistic imputation approaches across all datasets with different observed ratios are shown in table \ref{table:nllk}. We can see that \ours, an online method that only processes data once, beats the offline baselines and achieves the best performance in all cases. 

To further show the sensitivity of \ours over the different latent space dimensions, we evaluate our method on the Guangzhou-traffic dataset (observed ratio = 70$\%$) with different settings of latent space dimension and evaluate the test CRPC score. The results are shown in table \ref{table:dim-test}. We found that increasing the latent space dimension, especially the trend factor dimension ($D_r$), can improve the model performance. However, the improvement is not linear, and the model may suffer from overfitting when the latent space dimension is too large. A seasonal factor dimension ($D_s$) that is too high may also lead to overfitting and degrading the model performance.

\begin{figure*}
		\centering
		\setlength{\tabcolsep}{0pt}
		\begin{tabular}[c]{cc}
			\begin{subfigure}[b]{0.45\textwidth}
				\centering
				\includegraphics[width=\linewidth]{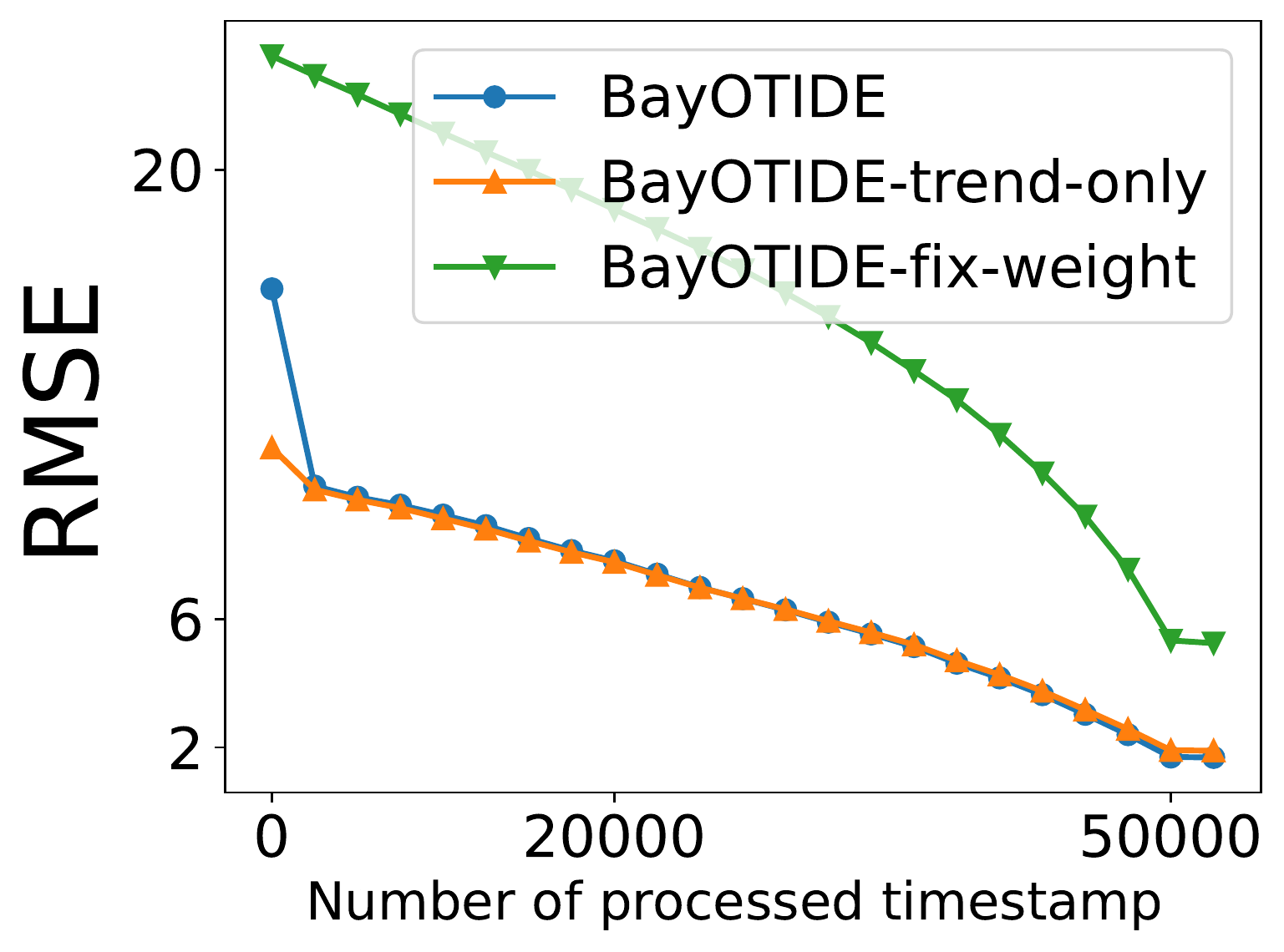}
                \caption{\small Solar-Power}
                \label{fig:running_rmse-solar}
			\end{subfigure} &
		\begin{subfigure}[b]{0.40\textwidth}
			\centering
			\includegraphics[width=\linewidth]{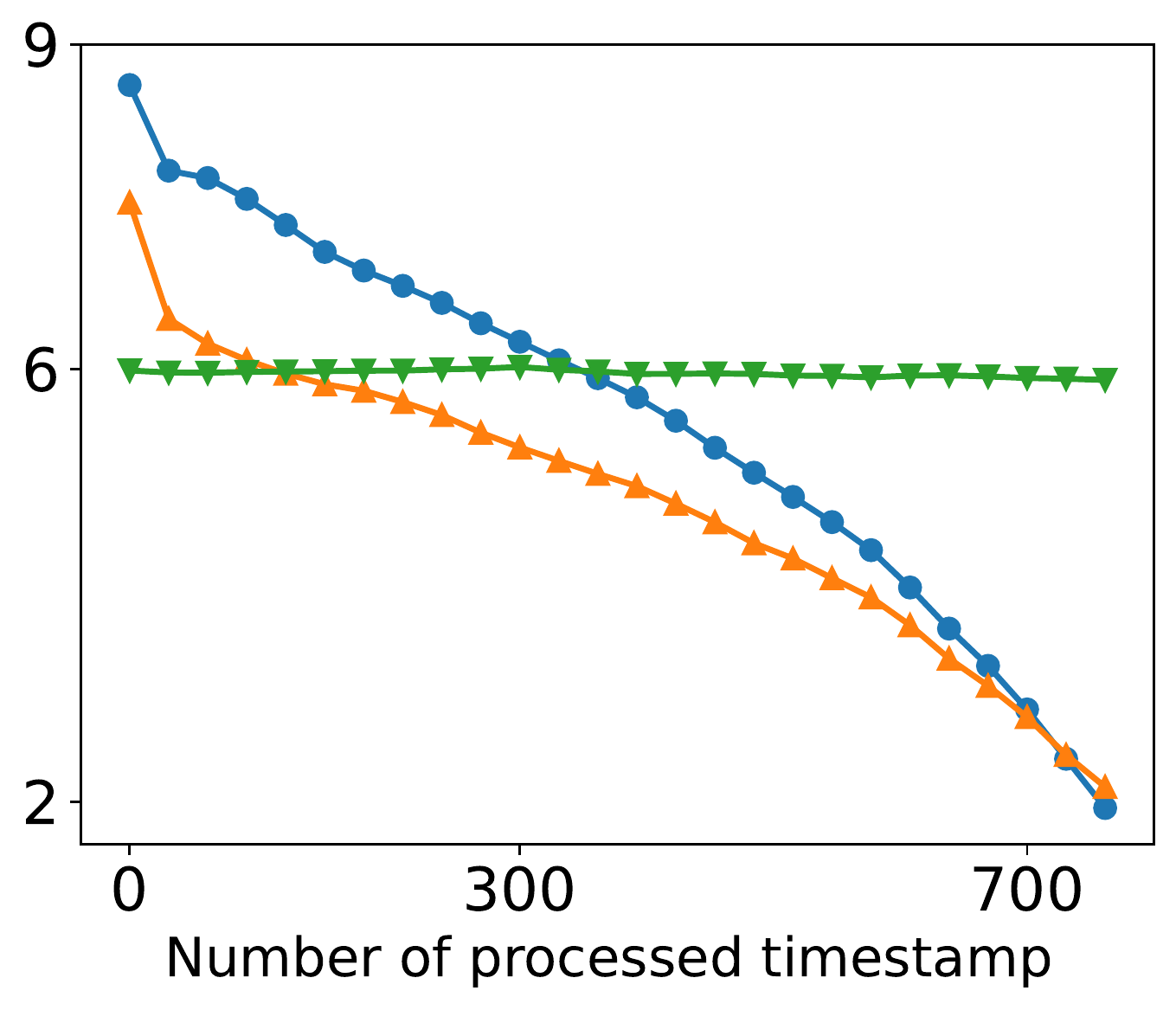}
            \caption{\small Uber-Move}
            \label{fig:running_rmse-uber}
		\end{subfigure} 
		\end{tabular}
		\vspace{-0.15in}
		\caption{\small Online imputation results  on \textit{Solar-Power} and \textit{Uber-Move}.}
		\label{fig:running_rmse_appendix}
	\end{figure*}

For the online imputation, the results on the \textit{Solar-Power} and \textit{Uber-Move} is shown in Figure \ref{fig:running_rmse-solar} and Figure \ref{fig:running_rmse-uber}.

\begin{table*}[]
	\centering
	\begin{small}
		\begin{tabular}{lcc|cc|cc}
			\toprule
                 {Dataset} & \multicolumn{2}{c}{Traffic-GuangZhou} &  \multicolumn{2}{c}{Solar-Power} &  \multicolumn{2}{c}{Uber-Move} \\
                 {Observed ratio}& $50 \%$ &  $70 \%$ & $50 \%$ &  $70 \%$ & $50 \%$ &  $70 \%$ \\ 
                \midrule
                  \multicolumn{6}{l}{\textit{Probabilistic \& Offline}}\\
                \midrule
                {Multi-Task GP} &{$7.339$} &{$6.921$} &{$4.921$} &{$4.292$}  &{$4.426$} &{$4.027$} \\
                {GP-VAE} &{$5.353$} &{$4.691$} &{$6.921$} &{$6.006$}  &{$7.323$} &{$5.827$} \\
                {CSDI} &{$3.942$} &{$3.518$} &{$3.433$} &{$2.921$}  &{$2.415$} &{$2.322$} \\
                {CSBI} &{$3.912$} &{$3.527$} &{$3.537$} &{$3.016$}  &{$2.424$} &{$2.331$} \\
                \midrule
                  \multicolumn{6}{l}{\textit{Probabilistic \& Online}}\\
                \midrule
                {BayOTIDE-fix weight} &{10.239} &{8.905} &{4.116} &{4.093}  &{3.249} &{3.252} \\
                {BayOTIDE-trend only} &\boldmath{$2.897$} &\boldmath{$2.852$} &{$1.944$} &{$1.878$}  &{$2.169$} &{$2.146$} \\
                {BayOTIDE} &{$3.244$} &{$3.078$}&\boldmath{$1.885$} &\boldmath{$1.852$}  &\boldmath{$2.167$} &\boldmath{$2.100$} \\
			\bottomrule
		\end{tabular}
	\end{small}
	\caption{\small The negative log-likelihood score (NLLK) of all probabilistic imputation methods on all datasets with observed ratio $=\{50\%,70\%\}$}
	\label{table:nllk}
\end{table*}

\begin{table*}[]
	\centering
		\begin{tabular}{lccccc}
			\toprule
                 {Test CRPC} & $D_r$ = 5	& $D_r$ = 10 & $D_r$ = 20 &	$D_r$ = 30	& $D_r$ = 40 \\
                 \midrule
                 $D_s= 3$ & {$0.068$}	& {$0.064$}& {$0.061$} &	{$0.057$}	& {$0.056$} \\
                 $D_s= 5$ & {$0.067$}	& {$0.063$}& {$0.059$} &	{$0.056$}	& {$0.055$} \\
                 $D_s= 10$ & {$0.066$}	& {$0.063$}& {$0.058$} &	{$0.053$}	& {$0.053$} \\
                 $D_s= 20$ & {$0.081$}	& {$0.074$}& {$0.065$} &	{$0.059$}	& {$0.058$} \\                 
			\bottomrule
		\end{tabular}
	\caption{\small The test CRPC on Guangzhou-traffic dataset(observed ratio = 70$\%$) with different settings of latent space dimension.}
	\label{table:dim-test}
\end{table*}

\end{document}